\documentclass[lettersize,journal]{IEEEtran}
\usepackage{amsmath,amsfonts}
\usepackage{algorithmic}
\usepackage{algorithm}
\usepackage{array}
\usepackage[caption=false,font=normalsize,labelfont=sf,textfont=sf]{subfig}
\usepackage{textcomp}
\usepackage{stfloats}
\usepackage{url}
\usepackage{hyperref}
\usepackage{verbatim}
\usepackage{graphicx}
\usepackage{nomencl}
\usepackage{cite}
\usepackage{bm}
\usepackage{ifpdf}
\usepackage{xcolor}
\usepackage[normalem]{ulem}

\newcommand{\bluer}[1]{\textcolor{black}{#1}}
\newcommand{\RomanNumeralCaps}[1]{\MakeUppercase{\romannumeral #1}}

\newcommand{\quotes}[1]{``#1''}
\makenomenclature

\graphicspath{{./figs/}}

\begin{document}

\title{Challenges and Opportunities in Deep Reinforcement Learning with Graph Neural Networks: A Comprehensive Review of Algorithms and Applications}




\author{\IEEEauthorblockN{Sai Munikoti,~\textit{Student Member, IEEE}, Deepesh Agarwal, Laya Das,\\ Mahantesh Halappanavar,~\textit{Senior Member, IEEE}, Balasubramaniam Natarajan,~\textit{Senior Member, IEEE} 
		\thanks{S. Munikoti and M. Halappanavar are with Data Science and Machine Intelligence group, 99354, PNNL, Richland, USA. (e-mail: sai.munikoti@pnnl.gov/ mahantesh.halappanavar@pnnl.gov )}
		\thanks{D. Agarwal and B. Natarajan are with Electrical and Computer Engineering, Kansas State University, Manhattan, KS-66506, USA, (e-mail: saimunikoti@ksu.edu, deepesh@ksu.edu, bala@ksu.edu)}
		\thanks{L. Das is with Reliability and Risk Engineering Lab, ETH Zurich, 8092, Zurich, Switzerland. (e-mail: laydas@ethz.ch )}
			\thanks{This work has been submitted to the IEEE for possible publication. Copyright may be transferred without notice, after which this version may no longer be accessible.}
		}}

\maketitle

\begin{abstract}
Deep reinforcement learning (DRL) has empowered a variety of artificial intelligence fields, including pattern recognition, robotics, recommendation-systems, and gaming. Similarly, graph neural networks (GNN) have also demonstrated their superior performance in supervised learning for graph-structured data. In recent times, the fusion of GNN with DRL for graph-structured environments has attracted a lot of attention. This paper provides a comprehensive review of these hybrid works. These works can be classified into two categories: (1) algorithmic \bluer{contributions}, where DRL and GNN complement each other \bluer{with an objective of addressing each other's shortcomings in a theoretical setting}; (2) application-specific \bluer{contributions}, \bluer{that leverage a combined GNN-DRL formulation to address problems specific to different applications}. This fusion effectively addresses various complex problems in engineering and life sciences. Based on the review, we further analyze the applicability and benefits of fusing these two domains, especially in terms of increasing generalizability and reducing computational complexity. Finally, the key challenges in integrating DRL and GNN, and potential future research directions are highlighted, which will be of interest to the broader machine learning community.  
\end{abstract}


\section{Introduction}
In the recent past, deep learning has witnessed an explosive growth in terms of development of novel architectures, algorithms and frameworks for addressing a wide range of challenging real-life problems ranging from computer vision to modeling to control. Among these developments, the use of deep neural networks (DNN) for solving sequential decision making problems within the reinforcement learning (RL) framework, resulting in deep reinforcement learning (DRL) is considered one of the state-of-the-art frameworks in artificial intelligence \cite{MIT2017}. 
This approach finds applications in combinatorial optimization \cite{mazyavkina2021reinforcement}, games \cite{mnih2013playing}, robotics \cite{finn2016guided}, natural language processing \cite{luketina2019survey}, and computer vision \cite{bernstein2018reinforcement}. The tremendous success of DRL in these applications can be credited to (1) the ability to tackle complex problems in a computationally efficient, scalable and flexible manner, which is otherwise numerically intractable \cite{farazi2021deep}; (2) high computational efficiency allowing fast generation of high fidelity solutions that are crucial in highly dynamic environments with demand for real-time decisions \cite{zhou2021deep}; (3) the ability to understand environment dynamics and produce near-optimal actions based solely on interactions with the environment, without the need for explicit prior knowledge of the underlying system \cite{frikha2021reinforcement, boute2021deep}. 


While DRL's effectiveness has most popularly been demonstrated in games, it is rapidly being adopted in various other real-life applications. Several of these applications involve environments exhibiting explicit structural relationships that can be represented as graphs \bluer{\cite{li2022confluence}}. For example, a network of cities in the Travelling Salesman Problem (TSP) or an incomplete knowledge graph are inherently characterized by a graph-based arrangement of the different entities. Methods developed for handling data in the Euclidean space are not well-suited for such environments, that require special treatment in terms of encoding the nodes or aggregating the information from different agents. These aspects are systematically modelled with graph neural networks \bluer{(GNNs)}, detailed in Section \RomanNumeralCaps{2}. Incorporation of such structural relationship serves as an auxiliary input, and further improves the quality of solutions. 

Recently, researchers have been exploring the idea of fusing powerful GNN models with DRL to efficiently tackle such graph-structured applications. \bluer{Although there are several advantages of this hybrid approach, there are two primary reasons associated with it. Firstly, DRL is rapidly being adopted in various applications that involve environments and formulations exhibiting structural relationships. GNNs are very effective in capturing such relationships and thus improves the naive model performance. More importantly, GNNs offer a computationally efficient and powerful framework for large-scale complex DRL environments, which is not feasible with other modeling paradigms. Secondly, there are several optimization-related tasks in GNN which can be efficiently handled with DRL compared to any other paradigm. For instance, tweaking graphs (by adding nodes and edges) in order to obtain robust models against adversarial attacks.} Therefore, a thorough review of these hybrid works could be extremely beneficial in identifying challenges and determining future research directions.\\
\textbf{Related work:}
\bluer{Several review works either related to DRL or GNN in general, are continuously being published \cite{mazyavkina2021reinforcement, luketina2019survey, bernstein2018reinforcement, farazi2021deep, frikha2021reinforcement, zhou2021deep, boute2021deep, perera2021applications, obite2021overview, glavic2019deep, botvinick2020deep, alomari2022deep,zhou2020graph,zhou2022graph,wang2017knowledge}. Most of these review articles include only a small paragraph or section that briefly presents the combination of the two approaches. For instance, \cite{wu2020comprehensive} explains various type of GNN architectures and their specific applications, but it discuss nothing about the DRL other than the fact that it can be used to generate graphs of desirable properties. Similarly, in \cite{ji2021survey}, RL is described briefly in the context of path finding and relation extractor for the knowledge graph in order to reason. DRL is primarily presented as a process or algorithmic steps, and the review lacks a comprehensive understanding of the applicability and significance of the particular DRL formulation.} In a nutshell, majority of these surveys are conducted via the lens of a particular application domain. As a result, they are confined to specific approaches that ignore holistic perspectives across domains. To the best of our knowledge, comprehensive reviews dedicated to the study of the combined potential of DRL and GNN do not exist in the current literature.  \\
\textbf{Contributions:} This paper focuses on a systematic literature review of the fusion of DRL and GNN, and makes the following contributions:
\begin{itemize}
    \item A rigorous review of articles spanning theoretical developments \bluer{(Section \RomanNumeralCaps{3}-A)} and multiple application domains \bluer{(Section \RomanNumeralCaps{3}-B)} \bluer{is conducted for hybrid DRL and GNN paradigm, which is rapidly gaining momentum and proving to be useful in several applications}. 
    \item A categorization of theoretical and application-specific contributions of the integrated DRL-GNN efforts is developed \bluer{(Section \RomanNumeralCaps{3})}. To this end, various attributes are identified for classifying and analyzing existing works \bluer{(Section \RomanNumeralCaps{4})}. 
    \item The survey takes a holistic approach to reviewing the literature with special focus on critical aspects of algorithms such as computational efficiency, scalability, generalizability and applicability. 
    \item Both DRL and GNN are still in early stages of development, as is the study of their fusion. Therefore, a thorough investigation of the associated challenges is performed and future research directions are identified, \bluer{which can advance the state-of-the-art both in theoretical and practical aspects} \bluer{(Section \RomanNumeralCaps{5})}.
\end{itemize}

This review is limited to articles indexed in IEEE Xplore, Scopus and Google Scholar. Initially, the keywords \quotes{deep reinforcement} and \quotes{graph neural network} are used to select articles from databases. This search led to more than $100$ papers from the year 2017 to 2022. The resulting list is filtered to identify articles that include both DRL and GNN which finally led to $40$ papers. \bluer{Most of the papers in this hybrid paradigm appear in a past few years, which } indicates the relatively recent history and relevance of this trending research topic.
Among the $40$ articles, $22$ come from conferences proceedings, $8$ from journals, and the remaining $10$ are preprint manuscripts.

The paper is organized as follows. In \bluer{(Section \RomanNumeralCaps{2})}, we offer a brief methodological background of both DRL and GNN to equip readers to understand the fundamentals prior to looking at the fusion of those techniques. \bluer{(Section \RomanNumeralCaps{3})} presents a comprehensive review of existing literature, including classification based on different novel attributes. In \bluer{(Section \RomanNumeralCaps{4})}, we discuss our findings in terms of the applicability and unique offerings of an approach involving GNNs and DRL. \bluer{(Section \RomanNumeralCaps{5})} highlights key limitations in the existing literature as well as potential future directions for research. \bluer{(Section \RomanNumeralCaps{6})} concludes this study.

\section{Overview of DRL and GNN}
\label{sec:overview}

\begin{figure*}[h!]
	\centering
	\includegraphics[width=\textwidth]{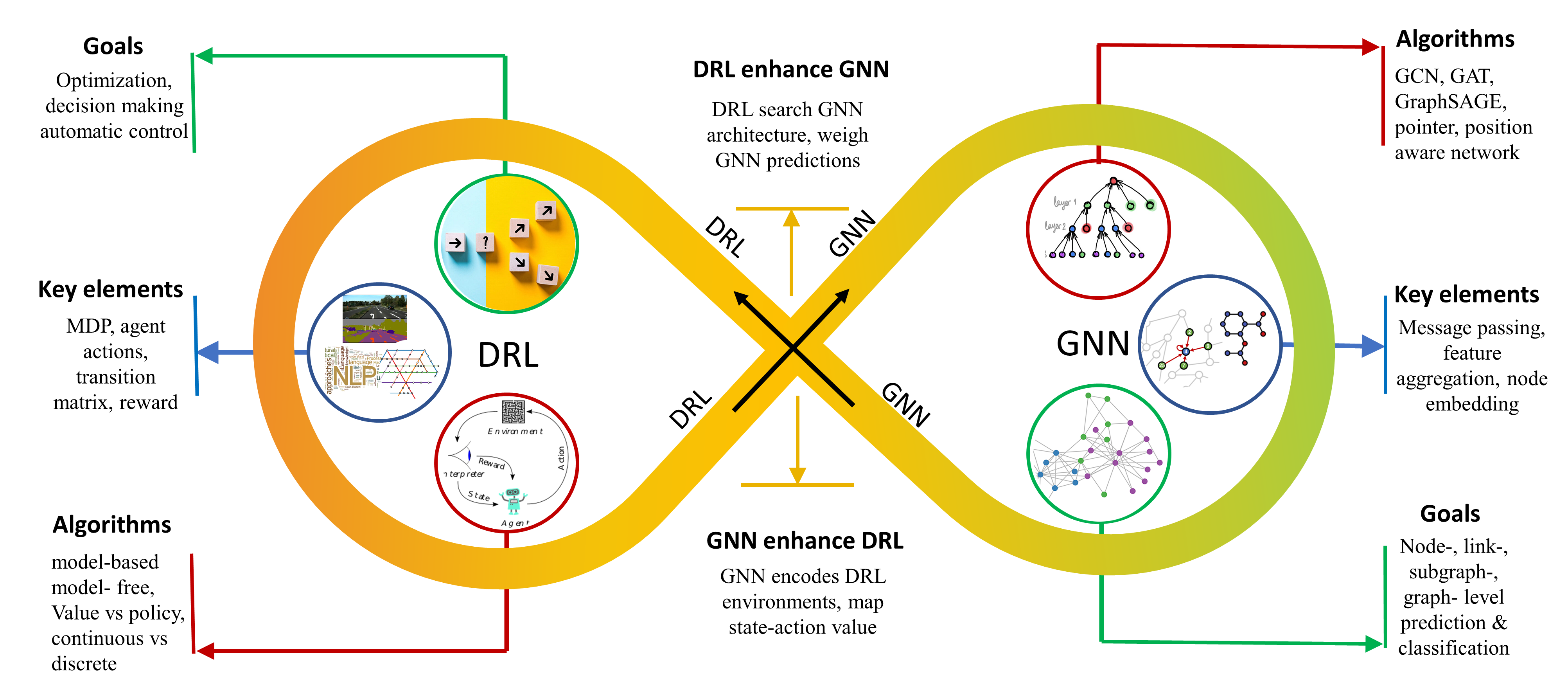}
	\caption{\bluer{Fusion of Deep reinforcement learning with Graph neural network}}
	\label{fig:framework1}
\end{figure*}

This section \bluer{begins with the definition of a graph, a mathematical model best suited to describe networked systems, and} provides the foundations of two powerful learning paradigms namely DRL and GNN. We \bluer{introduce } RL and \bluer{then present its extension with deep neural networks, i.e., }DRL. Then, we briefly explain the fundamentals of GNN algorithms. \bluer{ Fig. \ref{fig:framework1} depicts these two learning paradigms through an infinite shape loop. The left loop shows Deep Reinforcement Learning (DRL) paradigm, and the right loop exhibit the framework of Graph Neural Network (GNN). Through Fig. \ref{fig:framework1}, we first illustrate the contrasting view of these two frameworks in general by leveraging three attributes, namely \quotes{Goals}, \quotes{key steps} and \quotes{Algorithms}. \quotes{Goals} primarily represents the objectives or tasks that are typically accomplished with the framework. \quotes{Key-steps} describe the process involve in obtaining the algorithm. \quotes{Algorithm} highlights the widely known architectures developed under the framework. We also marked the intersection area of two loops, where the two paradigms come together in a fundamentally holistic way to solve complex problem in an elegant way.} This section will equip the readers with the required background knowledge to follow the hybrid works on DRL and GNN (discussed in \bluer{(Section \RomanNumeralCaps{3})}. 

\bluer{A graph $G$ is a mathematical model for representing a networked system. It is typically denoted via a tuple $G=(\mathcal{V},\mathcal{E})$ of a set of $n_V$ nodes $\mathcal{V}$ and $n_E$ links $\mathcal{E}$. The nodes represent entities of the underlying networked system (e.g. users in an online social network) and links denotes their relationship (e.g. friendship). Each node $u_i$, $i\in[1,n_V]$ of the graph may consist of features, typically represented through a vector $h_i$. Similarly, the link $e_{ij}$, between $u_i$ and $u_j$ could also be associated with a set of features represented as $h_{ij}$ which can, for example, signify the strength of the link. The interconnections in the graph can also be represented with an adjacency matrix $A\in\mathbb{R}^{n_V\times n_V}$ such that $A(i,j)=1$ if there exists a link between $u_i$ and $u_j$, and $0$ otherwise.}

\subsection{Deep Reinforcement Learning}

Reinforcement learning (RL) is considered the third important branch of machine learning with supervised and unsupervised learning serving as the other two \bluer{\cite{sutton2018reinforcement}}. RL is a sequential decision process where agents are trained to take optimal actions for different scenarios of an environment. The action transitions the environment to a new state and meanwhile the agent gets some reward that quantifies how good or bad the action was. To formulate the sequential decision process, RL employs a well known mathematical concept of Markov decision process (MDP) \bluer{\cite{sutton2018reinforcement}}. Typically, an MDP is defined by ($X,A,p,R$) where $X$ is a finite state space, $A$ is the action space for each state $x \in X$, $p$ is the state transition probability from state $x^{t}$ at time $t$ to state $x^{t+1}$ at time $t+1$, and $r$ is the immediate reward value obtained after an action $a \in A$ is performed. 
The agent’s primary goal is to interact with its environment (take state as input) at each time step to find the optimal policy $\pi^{*}$ (return action for the current state) in order to reach the goal while maximizing the cumulative rewards (expected return) over the entire time period. Agent takes the state $X$ as input and returns an action $a$ to be taken. At a particular time step $t$, the expected return $R^{t}$ is the sum of rewards from the current time step onward till the last time step $T$.
When taking an action, an agent must choose between taking the best action based on previous experiences (exploit) and gathering new experiences (exploration) in order to make better decisions in the future. A common approach to account for the trade-off is the greedy strategy, where the agent takes a random action with a probability $\epsilon$.

In addition, there are real-life situations where agents lack sufficient knowledge about the environment for holistic learning. Therefore, \textit{partially observed MDPs (POMDP)} is designed for these conditions \bluer{\cite{sutton2018reinforcement}}. \bluer{A POMDP is an MDP where the agent only possess a partial view of the state and therefore policy functions maps history of observations (belief states) to actions. It's typical expression is similar to that of MDP ($X,a, \Omega, T,p,O,r, b_{o}$) with some extra elements. The new elements include $\Omega$ that denotes observation, $T$ signifies time, $O$ represents observation probabilities, and $b_{o}$ is the initial probability distribution of states. Every time the agent takes the action, it receives an observation $ o \in \Omega$  which depends on the new state of the environment, the just taken action and probability $O (o| X^{'},a)$.}

\bluer{Traditional RL records the state, action and reward values for different actions taken at the time of training in a tabular format. However, this tabular approach is not scalable to larger number of state-action pairs, or to a continuous state-action space \cite{li2017deep}. This challenge can be addressed by employing a DNN to approximate the state-action values, leading to the paradigm of DRL. The DNN in a DRL framework estimates the state values and avoids a tedious record-keeping, thus providing an elegant solution to the scalability challenge. The added advantage of using DNN in DRL is that it allows the agent to generalize the value of states it has never seen before, or has partial information about, by leveraging the values of similar states. As a result, DRL algorithms are far more generalizable and practical for use across a wide range of applications involving vast state spaces.}

There are several ways to classify existing DRL algorithms such as: model-free vs model-based, value vs policy based, and offline vs online learning. In the following, we will provide the fundamental concepts of these algorithms across different categories.
\subsubsection{Value based DRL}
Value-based methods aim to learn the value of the state or state-action pair and then select actions accordingly. The state-action value function $Q_{\pi}(x,a)$, expressed in Eq. (\ref{eqn:2}), is the expected return starting from state $x$, taking action $a$, and thereafter following a policy {$\pi$}. {\em Deep Q learning} (DQN) is one of the widely use algorithm in this category \cite{mnih2015human}. Q-learning enables the agent to choose an action $a \in A$ with the highest Q-value available from state $x \in X$ based on a DNN model which maps discrete state-action space with Q values. \bluer{The parameters of
DQN network} is updated every time step following the Bellman optimality equation as shown in Eq. (\ref{eqn:2}), where $R$ is the reward obtained and $\alpha$ is the learning rate which takes values between $0$ and $1$. DQN is an ``off-policy'' algorithm, where a target policy is used to take action at the current state $X$ and a different behavior policy is used to select action at the next state.
\begin{equation}
\begin{split}
Q_{\pi}(x,a) = E_{\pi} (R^{t}|x^{t},a^{t}) = E_{\pi}(\sum_{k=0}^{\infty}\gamma^{k}r^{t+k}|x^{t},a^{t} ) \\
=(1-\alpha)Q(x,a) + \alpha \left[R + \gamma max_{a^{'} in A} Q(x^{t+1}, a^{t+1})\right]
\end{split}
\label{eqn:2}
\end{equation}
A key feature of DQN training is the replay buffer, which stores trajectory information $(x^{t}, a^{t}, r^{t}, x^{t+1})$ during each step of the training. In DQN, DNN is trained using a minibatch of a randomly selected sample (experiences) from replay buffer which offers various advantages in terms of sample efficiency, low variance and large learning scope. For each sample, the input (state) is passed through the current DNN to generate an output $\hat{Q}(x,a; \theta)$. The target $Q$ value corresponds to Bellman optimality equation in (\ref{eqn:2}), and is used to minimize the following loss function \bluer{\cite{sutton2018reinforcement}}:
\begin{equation}
L(\theta) = E[R + \gamma max_{a^{'} \in A} Q(x^{t+1},a^{'}) - Q(x,a;\theta)]
\label{eqn:4}
\end{equation}

DQN has many variations to improve its current design, including double DQN and dueling DQN. The max operator in the DQN update equation selects and evaluates an action using the same Q network. As a result, DQN significantly overestimates the value function. \textit{Double DQN} addresses this problem by employing two distinct networks, one for action selection and the other for action evaluation \bluer{\cite{van2016deep}}. Similarly, \textit{dueling Q} network approximates the Q function by decoupling the value function and the advantage function \bluer{\cite{wang2016dueling}}. 

\subsubsection{Policy based DRL}
These methods learn the policy directly unlike value based methods that learn the values first and then determine the optimal policy. Typically, a parametrized policy $\pi_{\theta}$ is chosen with parameters constantly updated by minimizing the expected return using a gradient based approach also known as {\em policy gradient theorem} \cite{sutton1999policy}. They are particularly suitable for very large action space (continuous problems) and learning stochastic policies. \bluer{The policy can be written as:
\begin{equation}
    \pi(a|x, \theta) = Pr( a | x, \theta),
    \label{eq:2a}
\end{equation}
which denotes the probability of taking action $a$ being at state $x$, and policy is parameterized by $\theta$. Now, we need an objective function to assess the performance of this policy. The objective function can be defined as:
\begin{equation}
    J(\theta) = v_{\pi_{\theta}}(S_{0}),
    \label{eq:2b}
\end{equation}
where $v_{\pi_{\theta}}(S_{0})$ is the true value function for the policy $\pi(a|x, \theta)$, and $S_{0}$ is the start state.  In short the maximizing $J(\theta)$ means maximizing $v_{\pi_{\theta}}(S_{0})$ which is equivalent to $\nabla J(\theta) = \nabla v_{\pi_{\theta}}(S_{0})$. According to policy gradient theorem,  
\begin{equation}
    J(\theta) \propto \sum_{x} \mu(x) \sum_{a}q_{\pi}(x,a)\nabla \pi(a|x,\theta),
    \label{eq:2c}
\end{equation}
where $\mu(x)$ is the distribution under $\pi$, $q(x,a)$ is the action value function, and $\nabla v_{\pi_{\theta}}(X_{0})$ is the gradient of $\pi$ given $x$ and $\theta$. So the theorem says that $J(\theta)$ is proportional to the sum of the q function times the gradient of the policies for all feasible actions at the states that we might be at. But to compute this gradient, we need to find $\nabla \pi(a|x,\theta)$. It turns out the gradient can be expressed as:
\begin{equation}
    \nabla_{\theta} \pi_{\theta}(x,a)= \pi_{}(x,a) \nabla_{\theta}Log \pi_{\theta}(x,a),
    \label{eq:2e}
\end{equation}
where $\nabla_{\theta}Log \pi_{\theta}(x,a)$ is scoring function. Since this is a gradient method, the update of the parameters (that we are trying to optimize) will be conducted in following way:
\begin{equation}
\Delta \theta = \alpha \nabla_{\theta}J(\theta).
    \label{eq:2d}
\end{equation}
There are various ways to define the policy such as softmax, Gaussian, etc. \cite{sutton2018reinforcement}
} Next, we discuss three widely used policy based methods:
($i$) \textit{REINFORCE}: where parameter updates at a given time step involves only the action taken from the current state -- the update relies on estimated return by Monte-Carlo method using episode samples. Since it relies on expected return from the current time step, it works only for the episodic tasks \cite{farazi2021deep};
($ii$) \textit{Trust region policy optimization (TRPO)} \bluer{\cite{schulman2015trust}}: \bluer{ add KL divergence constraints for enabling the trust-region for the optimisation process. It makes sure that the new updates policy is not far away from the old policy or we can say that the new policy is within the trust region of the old policy.} This constraint is in the policy space rather than in the parameter space. \bluer{The KL constraint adds additional overhead in the form of hard constraints to the optimisation process. Hence, there is a simpler approach to this problem in terms of  } 
($iii$) \textit{Proximal policy optimization (PPO)} \bluer{\cite{schulman2017proximal}}: that relies on clipped surrogate objective function to reduce the deviation between new policy and old policy. \bluer{Basically, it defines the probability ratio between the new policy and old policy, and led this ratio to stay within a small interval around $1$. The clip function truncates the policy ratio between the range  $1-\epsilon$  and $1+\epsilon$ , where $\epsilon$ is a hyperparameter. The objective function of PPO takes the minimum value between the original value and the clipped value. } It is relatively simpler in implementation and empirically performs on par with TRPO \cite{farazi2021deep}. \bluer{Furthermore, these algorithms (REINFORCE, TRPO, etc.) leverage value estimates for parameter updation. There is an another line of work which assume the presence of model information and uses techniques from dynamic programming and approximate dynamic programming to develop model aided reinforcement learning techniques such as actor only policy or direct controller \cite{pal2020brief}.}

\subsubsection{Actor-critic DRL}
Both value based and policy based algorithms have some limitations. While value based algorithms are not efficient for high dimensional action space, policy based algorithms have high variance in gradient estimates. To overcome these shortcomings, an actor-critic method has been proposed that combines the two approaches \cite{grondman2012survey}. 
Fundamentally, the agent is trained with two estimators. First, is an actor function that controls the agent's behavior by learning the optimal policy, i.e., provides the best action $a^{t}$ for any input state $X^{t}$. Second is a critic function that evaluates the action by computing the value function. 

Some of the popular variants of algorithms under this category are discussed next.
\textit{Advantage actor-critic (A2C)} consists of two DNNs -- one for actor and one for critic \cite{mnih2016asynchronous}. \bluer{The term \quotes{Advantage} corresponds to the temporal difference (TD) error or in other words its a prediction error. This error is the difference of the TD target (predicted value of all future rewards from the current state $x$) and the value of the current state evaluated from the critic network}. Besides A2C, \textit{asynchronous advantage actor-critic (A3C)} executes different agents in parallel on multiple instances of the environment instead of experience replay as in A2C \bluer{\cite{mnih2016asynchronous}}. Although A3C is memory efficient, its updates are not optimal as different agents work with different version of model parameters. \textit{Deep deterministic policy gradient (DDPG)} is an extension over deterministic policy gradient (DPG), which is designed for continuous action space \cite{lillicrap2015continuous}. DPG defines policy to be the function $\mu_{\theta}: X \rightarrow A$.
Here, instead of getting the integral over actions as in stochastic policy, we only need to sum over the state space as action is deterministic. 
DDPG employs a parameterized actor function with a parameterized critic function that approximates the value function using samples. In this way, DDPG can tackle large variance in policy
gradients of actor-only methods. 

\subsection{Graph Neural Network}
Learning with graph-structured data, such as knowledge graphs, biological, and social networks have recently attracted a lot of research attention. There are numerous benefits of representing data as graphs, such as systematic modeling of relationships, simplified representation of complex problems, etc. It is however challenging to interpret and evaluate such graph-structured data by employing conventional DNN-based learning methods. The fundamental mathematical procedures like convolutions are difficult to implement on graphs owing to their uneven structure, irregular size of unordered nodes and dynamic neighborhood composition. Graph Neural Networks (GNN) address these shortcomings by extending DNN techniques to graph-structured data \bluer{\cite{scarselli2008graph}}. GNN architectures can jointly model both structural information as well as node attributes. They provide significant performance improvement for graph-related downstream tasks like node classification, link prediction, community detection and graph classification \cite{madhawa2020active}. Typically, GNN models consist of a message passing scheme that propagates the feature information of the nodes to its neighbors until a stable equilibrium is reached. \bluer{ This process can mathematically be expressed as:
\begin{equation}
   h_u^{(l)} = f(h_v^{l-1}),v \in N(u) \cup u 
\end{equation}
where, $h$ is the node representing (embedding) vector and $N(.)$ denotes the neighborhood.
} Several GNN algorithms have been proposed to improve this message passing technique. We discuss some of the key approaches below.

\subsubsection{GCN}
Graph convolutional network \cite{kipf2016semi} is the first effort that \bluer{incorporate convolution operations (similar to CNN) on GNN.} The core idea behind any GNN is to generate unique Euclidean representation of nodes/links in the graph. Conventionally, spectral methods generate node representation vectors using eigen decomposition, but are computationally inefficient and are not generalizable.
GCN overcomes these challenges with its powerful approximation, where the update equation of the node representation vector $h_{u}$ at a particular layer $l$ is given by:
\begin{equation}
\begin{split}
h_{u}^{(l)} &= g\Bigg[\theta^{(l)} h_{u}^{(l-1)}A^{*}\Bigg] = \Bigg[ \theta^{(l)} h_{u}^{(l-1)}D^{-\frac{1}{2}}AD^{\frac{1}{2}}\Bigg],
\end{split}
\label{eq:GCN}
\end{equation}
where, $A$ is Adjacency matrix, $D$ is degree matrix, \bluer{ $\theta$ is the learnable parameter and $g$ is the activation function. $A^{*}=D^{-\frac{1}{2}}AD^{\frac{1}{2}}$} is normalized in this way to scale the node features and ensures numerical stability at the same time. It is important to note that GCN relies on the entire graph (i.e., full adjacency matrix) for learning node representation which is inefficient as the number of neighbors of a node can vary from one to thousands or even more and cannot be generalized to graphs of different sizes.   

\subsubsection{GraphSAGE}
It is an inductive node embedding approach that exploits node attributes to learn an embedding function \cite{hamilton2017inductive}. It supports simultaneous learning of topological structure as well as distribution of node features within a confined neighborhood. The fundamental premise is to train a neural network that can recognize structural properties of node neighborhood, thereby indicating its local role in the graph along with global position. Initially, the algorithm samples node features in the local neighborhood of each node in the graph-structured data. This is followed by learning appropriate functional mappings to aggregate the information received by each node as it propagates through the GNN layers. This inductive learning approach is scalable across graphs of different sizes as well as subgraphs within a given graph. The operation performed at $l$\textsuperscript{th} node embedding layer is given by: 
\begin{equation*}
\begin{split}
h_{u}^{(l)} &= f^{(l)}\left(h_{u}^{(l-1)},h_{N(u)}^{(l-1)}\right) =   g\Bigg[\theta_{C}^{(l)}h_{u}^{(l-1)} +
\theta_{A}^{(l)}\tilde{A}\Big(h_{N(u)}^{(l-1)}\Big)\Bigg]
\end{split}
\label{eq:graphsage}
\end{equation*}
where $\tilde{A}$ represents the aggregation operation; $g[\cdot]$ specifies the activation function; $h_{u}^{(l)}$ denotes the node embedding of node $u$ at $l$\textsuperscript{th} layer; $N(u)$ describes the neighborhood of node $u$; $\theta_{C}$ and $\theta_{A}$ are the parameters of the combination and aggregation operation of GNN, respectively. 

\subsubsection{GAT}
Graph attention network (GAT) assumes that contributions of neighboring nodes to the target node are neither predetermined like GCN nor identical like GraphSage. GAT adopts attention mechanisms to learn the relative weights between two connected nodes \bluer{\cite{velivckovic2017graph}}. The graph convolutional operation according
to single head GAT is defined as follows:
\bluer{
\begin{equation}
\begin{split}
h_{u}^{(l)} &=  g\Bigg[
\sum _{v \in N(u) \cup u} \alpha_{uv} \theta^{(l)}  h_{u}^{(l-1)}\Bigg] \\
a_{uv}^{l} & = \text{softmax}\Bigg(g\Bigg[a^{T}\Bigg(\theta^{(l)}h_{u}^{(l-1)}|| \theta^{l}h_{u}^{(l-1)}\Bigg)\Bigg]\Bigg),
\end{split}
\label{eq:gat}
\end{equation}
where the attention weight $\alpha_{uv} $ quantifies the connection strength between node $u$ and its neighbor $v$.} The attention weight is learned across all node-pairs using softmax function that ensures weights sum up to one over all neighbors of the node $u$. \bluer{For the case of Multihead attention, the update equation for the node embedding is expressed as:
\begin{equation}
\begin{split}
h_{u}^{(l)} &= \mathbin\Vert_{k=1}^{k} \Bigg[g\Big(
\sum _{v \in N(u) \cup u} \alpha_{uv}^{k} \theta^{(l),k}  h_{u}^{(l-1)}\Big)\Bigg]
\end{split}
\label{eq:gat2}
\end{equation}
where $\mathbin\Vert$ represents concatenation, $\alpha_{uv}^{k}$
 are normalized attention coefficients computed by the k-th
attention mechanism, and $\theta^{(l),k}$
is the corresponding linear transformation’s weight matrix.}
This mechanism selectively aggregates the neighborhood contributions and suppresses minor structural details. 

\section{Categorization of DRL+GNN Methods}
\label{sec:classification}
\begin{figure*}
	\centering
	\includegraphics[width=\textwidth]{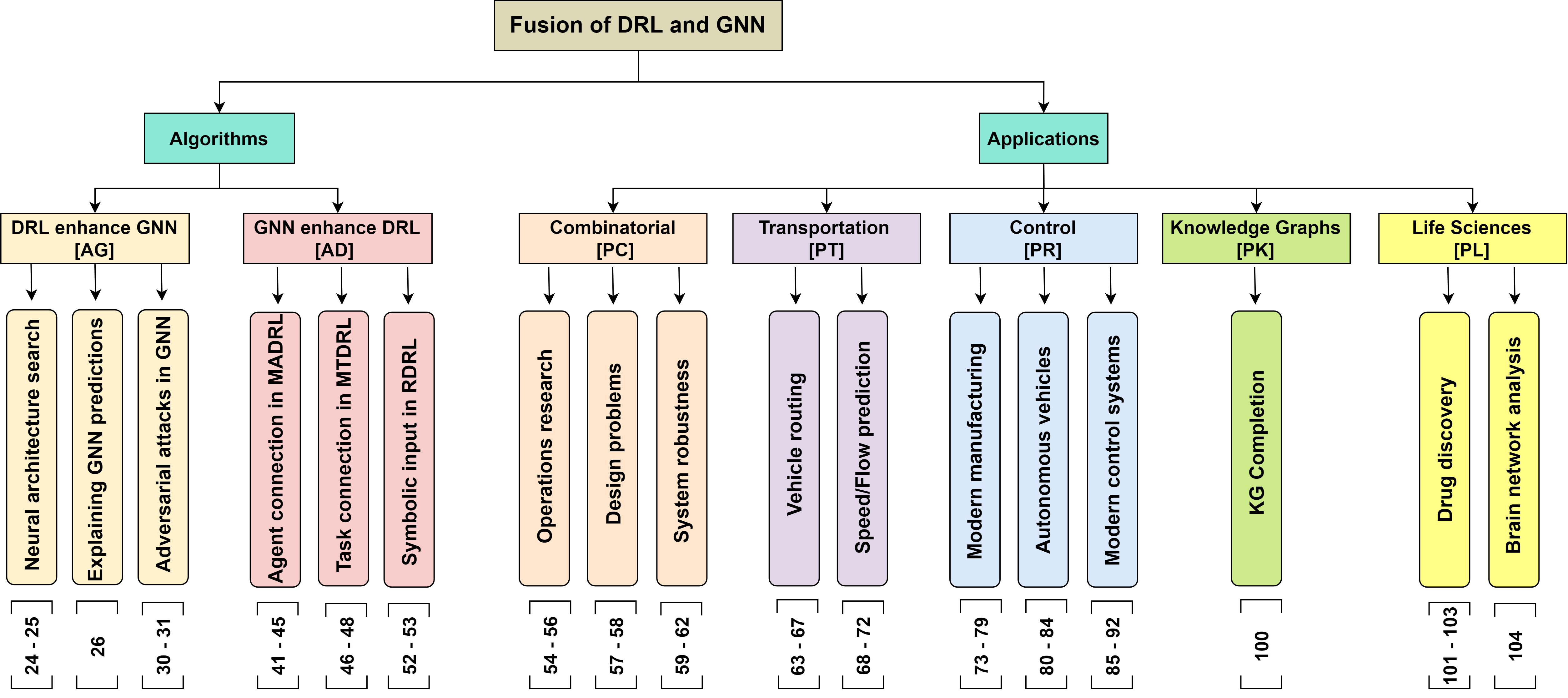}
	\caption{\bluer{A taxonomy of hybrid DRL and GNN works}}
	\label{fig:Taxonomy}
\end{figure*}

DRL and GNN have emerged as extremely powerful tools in modern deep learning. While DRL exploits the expressive power of DNNs to solve sequential decision making problems with RL, GNNs are novel architectures that are particularly suited to handle graph-structured data.
We identify two broad categories of research articles that jointly make use of GNN and DRL, as shown in Fig. \ref{fig:Taxonomy}. The first category of articles makes
\bluer{algorithmic contributions, where DRL and GNN complement each other with an objective of addressing each other's shortcomings in a theoretical setting.
On the other hand, the second category of articles makes
application-specific contributions, that leverage a combined GNN-DRL formulation to address problems specific to different applications. The first category involves novel fundamental formulations which enhance the performance of the algorithms in a generic way, irrespective of the application/use-cases. Whereas, papers in second category combined DRL and GNN to achieve a certain objective. Therefore, these works entail application-specific formulations that are suitable for particular use-cases.} The summary of surveyed DRL and GNN fused works is depicted in Table \ref{tab:Classification} and individual components of surveyed papers are outlined in Table \ref{tab:qualitativesummary}. \bluer{\quotes{Dynamic}, \quotes{Scalable}, \quotes{Generalizability} and \quotes{Multiagent} are four important attributes that are selected to characterize the existing works in Table \ref{tab:Classification}. Dynamic and Scalable are evaluated for GNN, where Dynamic determines whether the framework is applicable for time varying system (configuration, parameter), and Scalable indicates the ease (memory and computational) of building models for large-scale inputs and systems. On the other hand, Generalizability and Multi-agent are defined for DRL algorithms. Generalizability shows the applicability of trained model across similar RL environments. Multi-agent determines whether the DRL algorithm involves single agent or multiple agents to accomplish the desired task.}

\subsection{Algorithmic Developments}
\label{subsec:algorithms}
In this section, we discuss the articles that focus on developing novel formulations or algorithms to improve DRL or GNN. In these articles, either GNN is used to improve the formulation and performance of DRL, or DRL is used to improve the applicability of GNN.

\subsubsection{\textbf{DRL enhancing GNN}}
The articles that make use of DRL for improving GNN are used for diverse purposes including neural architecture search (NAS), improving the explainability of GNN predictions and designing adversarial examples for GNN. 

\textit{Neural Architecture Search (NAS):} refers to the process of automatically searching for an optimal architecture of a neural network (eg. number of layers, number of nodes in layer, etc.) to solve a particular task. A DRL-based controller that makes use of exploration guided with conservative exploitation is used in \cite{zhou2019auto} to perform an efficient search of different GNN architectures. The search space is made up of hidden dimension, attention head, attention, aggregation, combination, and activation functions. The authors introduce homogeneity of models as a method to perform a guided parameter sharing between offspring and ancestor architectures. \bluer{Along with computational gain, the unique characteristics of DRL based search engine is in terms of generalizability, i.e., same model can be used to quickly determine the optimal parameters for various use-cases. The superiority of the proposed method is demonstrated with better performance on benchmark datasets and baselines \cite{gao2020graph}}.

\textit{Explaining GNN predictions:} Generating explanations for DNN predictions is an important task in improving the transparency of ML models. Shan et al. \cite{shan2021reinforcement} use DRL to improve existing methods of explaining GNN predictions. The problem of generating explanations for GNN predictions involves identifying the sub-graph that is most influential in generating a prediction.
The authors devise a DRL-based iterative graph generator that starts with a seed node (the most important node for a prediction) and adds edges to generate the explanatory sub-graphs. \bluer{The reward of the DRL model is obtained as the mutual information
between the original predicted label and the label made by the generated graph. The distribution of predictions is based solely on the explanatory sub-graph to learn a sub-graph generation policy with policy gradient}. The authors show that the proposed method achieves better explainability in terms of qualitative and quantitative similarity between the generated sub-graphs and the ground truth explanations. \bluer{The unique exploitation vs exploration search mechanism of DRL inherently allow the model to find relevant subgraphs without getting stuck in local regions, which is usually the case with conventional approaches.}

\textit{Generating adversarial attacks for GNN:} Recent studies \cite{zugner2018adversarial, tang2020transferring, wu2019adversarial} have shown that GNNs are vulnerable to adversarial attacks that perturb or poison the data used for training them. DRL has been used to learn strategies to make adversarial attacks on GNNs, which in turn can be used to devise defense strategies to such attacks. 
RLS2V \cite{dai2018adversarial} is one of the first frameworks that uses DRL to perform an attack aimed at evading detection during classification. Specifically, it employs a Q-learning and structure-to-vector based attack methodology that learns to modify the graph structure (adding or dropping existing edges) with only the prediction feedback (reduction in accuracy) of the target classifier. \bluer{ While \cite{sun2020non} consider a novel node injection poisoning attack (NIPA) on graph data considering nodes as well as links.} Specifically, it injects fake nodes (e.g., fake accounts in a social networks) into the graph, and uses carefully crafted labels for the fake nodes together with links between them and other (fake as well as genuine) nodes in the graph to poison the graph data. NIPA frames the sequential addition of adversarial connections and the design of adversarial labels for the injected fake nodes as an MDP and solves it with deep Q-learning. To effectively cope with the large search space, NIPA adopts hierarchical Q-learning and 
GCN based encoding of the states into their low-dimensional latent representations to handle the non-linearity of the mapping between states and actions. \bluer{Furthermore, the underlying  Q network in these approaches, learns the sensitivity (relationship) between the changes in the input graph and final performance, rather than exact mapping. This enables them to predict performance for various possible configurations of attacks for different families of graph, which is not feasible with conventional graph optimization approaches. }

\subsubsection{\textbf{GNN enhancing DRL}}
This subsection discusses the papers related to the algorithmic improvement of DRL. Specifically, we focus on efforts wherein GNN has been used for relational DRL problems (RDRL) for effective modeling of the relationship among (1) different agents in a multi-agent deep reinforcement learning (MADRL) framework, and (2) different tasks in multi-task deep reinforcement learning (MTDRL) framework. 

\textit{Modeling relationship among agents in MADRL:} In MADRL, a group of agents cooperate or compete with each other to achieve a common goal. This framework has recently been used for a number of challenging tasks, including traffic light control, autonomous driving, and network packet delivery \cite{jin2018hierarchical, mao2019learning, mao2018modelling}. 
In such scenarios, the communication among agents offers additional information about the environment and state of other agents. Several methods have been proposed to learn this communication. 
The first body of work in capturing these relationships is related to attention-based approaches \cite{iqbal2019actor, su2020counterfactual, wang2020ahac, mao2020learning}. ATOC \cite{jiang2018learning}, DGN \cite{jiang2018graph}, and COMA-GAT \cite{su2020counterfactual} provide communication through the attention mechanism. 
Along these lines, G2ANet \cite{liu2020multi} employs hard attention to filter out irrelevant data and soft attention to focus on relevant information. DCG \cite{bohmer2020deep} employs coordination graphs, which uses message passing mechanism to coordinate the behaviors of agents. For each agent, these attention-based algorithms learn the significance distribution of other agents.
The authors in GraphComm \cite{shen2021graphcomm}, explore both static and dynamic relations simultaneously among agents. Specifically, it leverages the relational graph module to incorporate static relationships via relational graph provided by prior system knowledge and uses proximity relational graph for dynamic relationships. Agents' Q values are learned in a CTDE manner via MLP and GRU network along with RGCN and GAT to exchange messages among agents for static and dynamic relations, respectively.

Similarly, in \cite{zhang2021structural}, Zhang et al. propose a Structural Relational Inference Actor-Critic (SRI-AC) framework for CTDE that can automatically infer the pairwise interaction between agents and learn a state representation. The model is used to predict which agents need to interact in advance, and then supply the most relevant agent observation information to the critic network. In particular, each agent has a critic, which leverages information from the combined action as well as appropriate observational data during training. Then a variational autoencoder (VAE) is used to infer the pairwise interaction, and state representation from observed data, followed by a critic network that employs GAT to integrate the knowledge from neighboring agents. In a similar vein, \cite{yun2021multi} presents a novel state categorization method for CTDE DRL. Basically, it separates the state into agents' own observations, allies' partial information, and opponents' information specific to the Starcraft game, and then leverages GAT to learn the correlation and relationship among the agents.

\textit{Modeling relationship among tasks in MTDRL:} This framework provides an elegant way to exploit commonalities between multiple tasks in order to learn policies with improved returns, generalization, data efficiency, and robustness. One of the inherent assumptions in a majority of MTDRL works is compatible state-action spaces, i.e., same dimensions of states and actions across multiple tasks. However, this is violated in many practical applications like combinatorial optimization and robotics \bluer{\cite{mtdrl_01}}. This issue has been addressed by using GNNs that are capable of processing graphs of arbitrary size, thereby supporting MTDRL in incompatible state-action environments \cite{mtdrl_01}. Since GNNs provide the flexibility to incorporate structural information, it enables the integration of additional domain knowledge, where states are characterized as labeled graphs. The use of GNN in MTDRL has been demonstrated in continuous control environments that exploit physical morphology of the RL agents for constructing input graphs \cite{mtdrl_02, mtdrl_03}. Here, limb features are encoded in the form of node labels and edges represent the physical connections between the corresponding limbs. In this way, the structure of the agents is explicitly modeled in the form of graphs. NerveNet \cite{mtdrl_02} serves as the policy network which first propagates information over agent structure, followed by predicting actions for different parts of the agent. The authors in \cite{mtdrl_03} formulate a single global policy that can be represented as a collection of modular neural networks called Shared Modular Policies (SMP), each of which is designated to handle tasks related to its corresponding actuator.

\textit{Relational symbolic input for RDRL:} The fundamental premise of RDRL is to integrate DRL with relational learning or Inductive Logic Programming \cite{rdrl_01}, wherein states, actions and policies are represented by a first-order/relational language \cite{rdrl_02}. The tasks in this space are characterized by variable state and action spaces. In these problems, it is difficult to find fixed-length representation that is required by a majority of the existing DRL methodologies. This issue can be handled using GNN by formulating relational problems in terms of graph-structured data. The mechanics of a relational domain are typically represented by Relational Dynamic Influence Diagram Language (RDDL) \cite{rdrl_03}. Garg et al. \cite{rdrl_04} propose SymNet for automated extraction of objects, interactions, and action templates from RDDL. \bluer{Specifically, an RDDL problem instance is transformed into an instance graph, for which the embeddings of states and important object tuples are generated using GNN. Further, these are decoded into scores corresponding to individual ground actions.} Node embeddings are generated using GNN and action templates are applied over object tuples to create a probability distribution. This is followed by updating the model using policy gradient method. However, SymNet is computationally expensive and is applicable only when RDDL domain definition is available since pre-defined transition dynamics are required to construct the graphs. Symbolic Relational DRL (SR-SRL) \cite{rdrl_05} addresses these limitations by considering an enriched symbolic input comprising of objects and relations along with their features in the form of a graph. This does not require information about transition dynamics and generalizes over any number of object tuples.

\bluer{The unique aspect of GNN in terms of representing nodes/links in high dimensional embedding space makes them an powerful candidate for quantifying relationship among the large space of agents and tasks as described in previous three sub-sections. As a result, they are very successful in assimilating the relational information compared to any other relational learning methodologies.}

\bluer{In a nutshell, the review of papers in the part A (i.e., algorithmic developments) indicates that the first set of methods primarily addresses challenges associated with GNN models. Specifically, these challenges (such as explainability of GNN predictions and computational complexity of training and inference process) typically involve large optimization problems, which can be effectively handled with DRL. The unique \quotes{learning from interaction} approach makes DRL a suitable candidate for this kind of sequential identification problem. Furthermore, it exhibit all the desirable advantages of a learning paradigm, i.e., low computational effort, large scalability and high generalizability compared to the conventional Graph optimization methodologies. On the other hand, the second class of works focuses on improving the encoding aspects of DRL algorithms. Specifically, the complexity of the decision problems is continuously increasing where DRL is adapting to multi-agent and multi-task configurations. This demands a powerful encoding mechanism to represent the underlying systems and scenarios, which can be efficiently accomplished with GNN. The unique message passing mechanism combined with attention-based learning process, makes them an efficient encoder for complex data generated from distributed multi-agent and multi-task systems.}

\subsection{Applications}
\label{subsec:applications}
The second broad category of articles exploit the versatility of DRL along with the flexible encoding capability of GNNs to address interesting challenges in different application domains. These domains span a wide spectrum, including combinatorial optimization, transportation, control, knowledge graphs and life sciences, which we briefly review next.

\subsubsection{\textbf{Combinatorial optimization (CO)}} Many CO problems are computationally expensive and require approximations and heuristics to solve in polynomial time. There has been an increasing interest in solving CO problems using \bluer{DRL since they naturally fit into the problem}. In this regard, CO problems are often framed as an MDP where the optimal actions/solutions can be learnt with DRL. Further, the underlying environment is represented as graph that is processed using GNN. \bluer{ The problem can be mathematically written as: Given a  graph $G = (V, E)$, where $V = A \cup B$ and a budget $b$, find a set $S^{*} \subseteq A$ of $b$ nodes such that $f(S^{*})$ is maximized. $f(S^{*})$ is the cost function specific to the objective. In DRL, this problem is typically formulated as a MDP, where state represent the current solution set $S$ and action refers to the process of selecting a node from $V$ and appending to the solution set. The process is repeated until the cardinality $|S|$ reaches $b$. Reward quantifies the benefit of taking an action, and it is proportional to $f(S)$.  }
The articles addressing these challenges can further be divided into following sub-categories:

\textit{Solving CO problems in Operations Research:}
Manchanda et al.~\cite{manchanda2020gcomb} use GNN to capture the structural information of CO problems and thus address the poor generalizability and scalability of existing DRL-based approaches. 
They learned a construction heuristic for a budget-constrained maximum vertex cover (MVC) problem by combining supervised learning and DRL. 
GCN is first utilized to find appropriate candidate nodes by learning the scoring function computed using the probabilistic greedy approach. Then, the candidate nodes are used in an algorithm similar to \cite{khalil2017learning} to sequentially construct a solution. 
Since the degree of nodes in large graphs can be rather high, importance sampling based on computed score is used to select the neighboring nodes while determining embeddings, thereby reducing the computational complexity.
Extensive experiments on random/real-world graphs reveal that the proposed method marginally outperforms S2V-DQN and scales to much larger graph instances up to a hundred thousand nodes. Additionally, it is significantly more efficient in terms of the computation efficiency due to a relatively smaller number of learned parameters.

Another interesting application of DRL and GNN can be seen in solving diffusion processes in graphs, such as influence maximization, and epidemic test prioritization, among others. The goal is to identify a set of nodes on a temporally evolving graph such that the global objective of curbing spread or maximizing information spread is achieved \bluer{\cite{chen2021contingency,Munikoti2022GraMeR}}. Various graph-theoretic algorithms have been developed to address this class of problems. However, they are inefficient when scaling to larger graphs. Further, the added difficulty is that the states are partially observed, for instance, we might not know the ground truth infection status for every node in the graph at any point in time. To address these challenges, \cite{meirom2021controlling} poses the problem of controlling a diffusive process on a temporally evolving graph as a partially-observed Markov decision process (POMDP). The problem of selecting a subset of nodes for dynamical intervention is formulated as a ranking problem, and an actor-critic proximal policy optimization is employed to solve it. Specifically, the architecture contains two separate GCN modules; one updates the node representation according to the dynamic process and the other is in charge of long-range information propagation. The results on various real-world networks, including COVID-19 contact tracing data, show the superior performance of this approach.

\textit{Solving design problems}: Several design problems, especially electronic circuit design are CO problems, which can benefit from a DRL-GNN formulation. For instance, automatic transistor sizing is a challenging problem in circuit design due to the large design space, complex performance tradeoffs, and fast technological advancements. 
The authors in \cite{wang2020gcn}, present the GCN-RL Circuit Designer, which uses DRL to transfer knowledge between different technology nodes and topologies. GCN is employed to learn the circuit topology representation. The GCN-RL agent retrieves features of the topology graph with transistors as nodes and wires as links. Actor critic approach with continuous space DRL algorithm DDPG is used. The generalizability of DRL enables training on one technology node and then applying the trained agent to search the same circuit under different technology nodes. GCN extracts circuit features which enables transfer of knowledge between different topologies that share similar design principles, for example, between a two-stage and three-stage trans-impedance amplifier.

A similar problem is Logic synthesis for combinational circuits, in which the lowest equivalent representation for Boolean logic functions is sought. A widely used logic synthesis paradigm represents Boolean logic with standardized logic networks, such as and-inverter graphs (AIG), which iteratively conducts logic minimization operations on the graph. 
To this end, \cite{zhu2020exploring} poses this problem as an MDP and DRL is incorporated with GCN to explore the solution search space. Specifically, this work leverages Monte Carlo policy gradient based RL algorithm, REINFORCE. Since circuits and AIG logic can be naturally modeled as graphs, they leverage GCN to extract the current state’s features.

\textit{System robustness:} Recently, \bluer{authors in \cite{darvariu2021goal}} demonstrate a novel application of DRL with GNN, where the authors used DRL to search for optimal graph topology for a given graph objective. Essentially, the construction of a graph is framed as a sequential decision process of adding a fixed number of links to the current graph one at a time such that the robustness score of the final graph is the maximum among all feasible combinations of the given graph and edges. Particularly, the state represents the current graph, while the action corresponds to a new node that needs to be added. They are encoded using the S2V variant of GNN \cite{dai2016discriminative}, and DQN constitutes the underlying DRL engine. Although this approach is more computationally efficient than conventional approaches (such as greedy, Fiedler vector, etc.), it requires an iterative algorithm at each step of the episode to compute the global score of the current graph (robustness in this case), which demands some computational effort. This can be avoided by using learning-based models to compute intermediate rewards, i.e., global graph scores \cite{munikoti2022scalable, munikoti2021bayesian}.  

\subsubsection{\textbf{Transportation}}
Transportation problems that are handled with DRL and GNN can be broadly classified into two classes, namely routing and speed prediction. 

\textit{Vehicle routing:} \bluer{Routing problem is usually formulated by considering a graph $G(V , E)$, in which $V = C \cup \{0, n + 1\}$ is the set of nodes associated to customers in $C$ and to the depot nodes $0$ and $n + 1$. Nodes $0$ and $n+1$ represent the same depot and impose that all routes must start on $0$ and return to $n + 1$. Set $E$ contains the links $(u, v)$ for each pair of nodes $u , j \in N $. The cost of crossing an link $(u, v) \in E$ is denoted by $c_{uv}$ . Each node has a demand $q_{i}$. The objective of the problem is to determine a set of minimal cost route for each of the $k$ vehicle satisfying all the requirements. In DRL, this problem is typically formulated by initializing an solution set $S$ with the fixed starting and ending node representing depot. Action would be to select the city node and fill the intermediate values in the solution set $S$. Reward will be proportional to both the cost of the links ($c_{uv}$) and demand of the selected node $a$. The process is repeated until the aggregated demand of all the selected city crosses the net limit. This problem can also be posed in a multiagent setting where each agent (here each vehicle) allots its routes by exchanging the information among them.} One of the early efforts to solve the vehicle routing problems (VRP) with GNN with DRL is the Traveling Salesperson Problem (TSP), where the objective is to find the shortest possible route that visits each node in the graph exactly once and returns to the source node~\cite{deudon2018learning}. Here, the state is denoted by a graph embedding vector that describes the tour of the nodes until time step $t$, whereas action is defined as selecting a node from the non-visited pool and the reward is the negative tour length. A GNN with attention mechanism is used as an encoder followed by a pointer network decoder. The described encoder–decoder network's parameters are updated using the REINFORCE algorithm with a critic baseline.
The methodology proposed by \cite{drori2020learning} adopts a GNN representation to offer a general framework for model-free RL that adapts to different problem classes by altering the reward. This framework uses the edge-to-vertex line graph to model problems and then formulates them in a single-player game framework. The MDPs for TSP and VRP are the same as in \cite{bello2016neural}. Rather than employing a full-featured Neural MCTS, \cite{drori2020learning} represents a policy as a graph isomorphism network (GIN)  encoder with an attention-based decoder, which is learned throughout the tree-search operation. Further, \cite{lu2019learning} proposes to learn the improvement heuristics (methods that start from an arbitrary policy and improve iteratively) for VRP in a hierarchical manner. The authors devised an intrinsic MDP that includes not only the present solution's features but also the running history. REINFORCE method is used to train the policy, which is parameterized by a GAT.

Another important problem of cooperative combinatorial optimization in TSP is related to optimization of the multiple TSPs (MTSP). 
\cite{hu2020reinforcement} developed an architecture consisting of a shared GNN and distributed policy networks to learn a common policy representation to produce near-optimal solutions for the MTSP. Specifically, Hu et al. use a two-stage approach, where REINFORCE is used to learn an allocation of agents to vertices, and a regular optimization method is used to solve the single-agent TSP associated with each agent.

\textit{\bluer{Speed/flow control}:} The second class of transportation problems deals with the prediction of speed/flow in the road network commonly referred as traffic signal control (TSC) \bluer{ \cite{srinivasan2006neural}}. In recent years, TSC has been modeled as an MDP and researchers have adopted DRL to control the traffic signals \cite{yang2019cooperative, devailly2021ig, yang2021ihg, yoon2021transferable,noaeen2022reinforcement}. 
The authors in \cite{yang2021ihg} propose Inductive Heterogeneous Graph Multi-agent Actor-critic (IHG-MA) algorithm consisting of three steps: ($i$) sampling of heterogeneous nodes via fast random walk with restart approach, ($ii$) encoding heterogeneous features of nodes in each group using Bi-GRUs, and ($iii$) aggregating embeddings of groups using graph attention mechanism. Finally, the proposed MA framework employs the actor-critic approach on the obtained node embeddings to compute the $Q$-value and policy for each SDRL agent, and optimizes the whole algorithm to learn the transferable traffic signal policies across different networks and traffic conditions.
Shang et al. \cite{shang2022new} proposed to use a DQN agent to effectively combine the predictions of GCN and GAT, thus improving the overall space-time modeling capabilities and forecasting performance. The DQN provides weights to combine the GCN and GAT predictions, where the weights are adaptive to different network topology, weather conditions, and other relevant attributes of the traffic data. \bluer{Apart from the road network, traffic is also monitored in communication domain such as optical network. In \cite{almasan2022deep}, the authors have proposed to use message passing GNN architecture to solve routing optimization in optical networks. The key contribution of this work lies in its generalizability as it is able to achieve outstanding performance over several real-world networks that are never seen during training.}


\subsubsection{\textbf{Manufacturing and control}}
%
DRL has also been explored in modern manufacturing systems because of the increasing complexity and interdependency across processes and system-levels \cite{xiao2019meta, dornheim2020model, huang2020deep}. 
Recently, Huang et al. \cite{huang2021integrated} proposed an integrated process system model based on GNN.
Here, the manufacturing system is represented as a graph, where machines are treated as nodes and material flow between machines as links. GCN is used to encode machine nodes and obtain a node's latent representation that reflects both the local condition of the machine (i.e., parameters of neighboring machines) and the global status of the entire system. Each machine is modeled as a distributed agent, and MARL is trained to learn an independent adaptive control policy conditioned on the node's latent feature vector. The latent characteristics of the node, machine process parameters, and total yield with defects serve as the state, action, and reward of the underlying MDP, respectively. Specifically, C-COMA \cite{su2020counterfactual} has been deployed by employing the Advantage Actor Critic (A2C) framework in a distributed setting and is easily compatible with GNN.

In manufacturing, job shop scheduling problem (JSSP) is also an important problem that aims to determine the optimal sequential assignments of
machines to multiple jobs consisting of series of operations while preserving the problem constraints. 
Park et al. \cite{park2021learning} propose a framework to construct the scheduling policy for JSSPs using GNN and DRL. They formulate the scheduling of a JSSP as a semidefinite programming problem (SDP) in a computationally efficient way by representing the state of a JSSP as a disjunctive graph \cite{yamada2003studies}, where nodes represent operations, conjunctive edges represent precedence/succeeding constraints between two nodes,
and disjunctive edges represent machine-sharing constraints between two operations. Then, they employ a GNN to learn node embeddings that summarize the spatial structure of the JSSP and derive a scheduling policy that maps the embedded node features to scheduling action. 
Proximal policy optimization (PPO) algorithm, a variant of policy-based RL, is used to train GNN-based state representation module and the parameterized decision-making policy jointly \cite{schulman2017proximal}.  

Another key application of DRL is in the control of connected autonomous vehicles \cite{chen2020deep, saxena2020driving, du2020cooperative,wang2021reinforcement}.  However, DRL-based controllers in most existing literature address only a single or fixed number of agents with both fixed-size observation and action spaces. 
This is because of the highly combinatorial and volatile nature of CAV networks with dynamically changing number of agents (vehicles) and the fast-growing joint action space associated with multi-agent driving tasks, which pose difficultly in achieving cooperative control. Recently, Chen et al. \cite{chen2021graph} presented a DRL based algorithm that combines GCN with DQN to achieve efficient information fusion from multiple resources. A centralized multi-agent controller is then built upon the fused information to make collaborative lane changing decisions for a dynamic number of CAVs within the CAV network. 

Efficient allocation of communication resources in wireless networks that are commonly used in modern control systems to exchange data across a vast number of plants, sensors, and actuators is also addressed with DRL 
\cite{baumann2018deep,lima2020ieeeresource}. These DRL techniques, however do not scale well with the network size. 
To overcome this issue, Lima et al. \cite{lima2020resource} employ a GNN to parameterize the resource allocation function. In particular, Gama et al. \cite{8579589} use random edge GNNs  since the underlying communication graph is randomly distributed, and then coupled it with REINFORCE since the action space is continuous.

Another interesting application can be found in multi agent formation control. Although a number of algorithms can achieve formation control effectively, they ignore the structure feature of the graph formed by agents \cite{lowe2017multi, foerster2018counterfactual,iqbal2019actor}. 
Wang et al. \cite{wang2020multi} proposed a model named MAFCOA building on the framework of GAT. In particular, the model can be divided into two parts including formation control and obstacles avoidance. 
The first part uses GAT and focuses on cooperation among agents, while the second part focuses on obstacle avoidance with multi-LSTM models. 
The Multi-LSTM allows the agents to take obstacles into consideration in the order of distance and avoid arbitrary number of obstacles \cite{everett2018motion}. Moreover, in order to scale to more agents, the parameters are shared to train all the agents in a decentralized framework. Actor and critic approach is used with MADDPG to learn the optimal control policy for multiple agents. 

\subsubsection{\textbf{Knowledge graph completion}}
Knowledge Graphs (KG) are increasingly being used to represent heterogeneous graph-structured data in a wide variety of applications including recommendation systems \cite{kg_appl_recSys}, social networks \cite{kg_appl_socNet}, question-answering systems \cite{kg_appl_questAns}, smart manufacturing \cite{kg_appl_smrtMnfg}, information extraction \cite{kg_appl_infoExtr}, semantic parsing \cite{kg_appl_semnPars} and named entity disambiguation \cite{kg_appl_ned}. 
One of the key problems in real-world knowledge bases is that they are notoriously incomplete, i.e., a lot of relationships are missing. KG Completion (KGC) is a knowledge base completion process that aims to fill-in the incomplete real-world knowledge bases by inferring missing entries with the help of existing ones. The entities and corresponding relations are represented by means of triplets consisting of head nodes ($h$), relations ($r$), and tail nodes ($t$). The problem of KGC entails prediction of missing tails for given pairs of head nodes and relations. 
Traditional RL-based methods do not consider generation of new subgraphs within existing knowledge graphs, e.g., new or missing target entities. Moreover, the domain-specific rules are not incorporated, which could be utilized to learn state transition processes when KGC is posed as a Markov process. Finally, the issue of reward sparsity leads to large variance of sampling methods and low learning efficiency. In order to overcome these limitations, \cite{kgc_reinLrn_03} proposed a \bluer{Generative Adversarial Network (GAN)}-based DRL framework (GRL). The authors divide the problem into two scenarios: when the target entity can be located within limited time steps, and when the target entity cannot be found from the original KG while there are still time steps to go, in which case a new sub-graph is formed. KGC being defined as an MDP, explores the rules that can be introduced in both the state transition process and rewards to better guide the walking path under the optimization of GAN. LSTM is employed as a generator of GAN, which not only records previous trajectories (of states, actions, etc.) but also generates new sub-graphs and trains policy networks with GAN. Furthermore, to better generate new sub-graphs, a GCN is used to embed the KG into low-dimensional vectors and parameterize the message passing process at each layer. In addition, GRL also applies domain-specific rules and utilizes DDPG to optimize rewards and adversarial loss.

\subsubsection{\textbf{Life sciences}}
Along with engineering applications, recent advancements in ML have also demonstrated the potential to revolutionize various life sciences applications such as drug discovery \cite{mcnaughton2022novo, knutson2021decoding, do2019graph} and brain network analysis \cite{zhao2022deep}.
To this end, \cite{mcnaughton2022novo} proposes a new method for designing antiviral candidates coupling DRL to a deep generative model. Specifically, the authors use the actor critic approach, in which a scaffold-based generative model is leveraged as the actor model to build valid 3D compounds. For the critic model, parallel GNNs are used as a binding probability predictor to determine whether the generated molecule actively binds with a target protein \cite{knutson2021decoding}. The results demonstrated that the model could produce molecules with higher druglikeness, synthetic accessibility, water solubility, and hyrdophilicity than current baselines.
Do et al. \cite{do2019graph} proposed a Graph Transformation Policy Network (GTPN) that combines the strengths of DRL and GNN to learn reactions directly from data with minimal chemical knowledge. 
Their model has three key components: a GNN, a node pair prediction network, and a policy network. The GNN is responsible for obtaining the atom's representation, the node pair prediction network is responsible for computing the most possible reaction atom pairs, and the policy network is responsible for determining the optimal sequence of bond changes that transforms the reactants into products. Additionally, the model's step-by-step creation of product molecules allows it to exhibit intermediate molecules, greatly improving its interpretability. 

\begin{table*}
	\scriptsize
	\centering
	\caption{Summary of surveyed DRL and GNN fused works. AD:Algorithms-DRL enhancing GNN, AG:Algorithms-GNN enhancing DRL, PC:Applications in Combinatorial optimization, PT:Applications in Transportation, PR:Applications in Control, PK:Applications in Knowledge graph, PL:Applications in Life science}
	\label{tab:Classification}
	\begin{tabular}{|p{2.3cm}|p{1.0cm}|p{0.8cm}|p{1.5cm}|p{1.5cm}|p{1.5cm}|p{1.5cm}|p{1.7cm}|}
		\hline
		\textbf{Reference} & \textbf{Category} & \textbf{Dynamic} & \textbf{Scalable} & \textbf{Generalizable across envs} & \textbf{Multiagent} &  \textbf{Source code} & \textbf{Publication venue-year} \\
		\hline
		\multicolumn{8}{|c|}{\bf Algorithms} \\ \hline
		\bluer{AGNN} \cite{zhou2019auto} & AD &
		& \hspace{0.5cm} \includegraphics[width=0.25cm, height= 0.25cm]{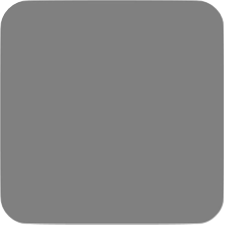}  & \hspace{0.50cm} \includegraphics[width=0.25cm, height= 0.25cm]{gray_images-rounded.png}  &   & & arxiv 2019 \\  
  \hline
    	\bluer{GraphNAS} \cite{gao2020graph} & AD &
		& \hspace{0.5cm} \includegraphics[width=0.25cm, height= 0.25cm]{gray_images-rounded.png} 
  & \hspace{0.50cm} \includegraphics[width=0.25cm, height= 0.25cm]{gray_images-rounded.png}  &  & \hspace{0.50cm} \href{https://github.com/GraphNAS/GraphNAS}{\includegraphics[width=0.25cm, height= 0.25cm]{gray_images-rounded.png}} & IJCAI 2020 \\
  \hline
  		\bluer{RG-Explainer} \cite{shan2021reinforcement} & AD &
		\hspace{0.20cm} \includegraphics[width=0.25cm, height= 0.25cm]{gray_images-rounded.png} 
		 & 
  & \hspace{0.50cm} \includegraphics[width=0.25cm, height= 0.25cm]{gray_images-rounded.png}  &   & & NeurIPS 2021 \\  
    \hline
	\bluer{RL-S2V} \cite{dai2018adversarial} & AD &
		& \hspace{0.5cm} \includegraphics[width=0.25cm, height= 0.25cm]{gray_images-rounded.png}  & 
		\hspace{0.50cm} \includegraphics[width=0.25cm, height= 0.25cm]{gray_images-rounded.png}  &   & \hspace{0.50cm} \href{https://github.com/Hanjun-Dai/graph_adversarial_attack}{\includegraphics[width=0.25cm, height= 0.25cm]{gray_images-rounded.png}} & ICML 2018 \\  
  \hline
  	Sun et al. \cite{sun2020non} & AD &
		& \hspace{0.5cm} \includegraphics[width=0.25cm, height= 0.25cm]{gray_images-rounded.png}
  & \hspace{0.50cm} \includegraphics[width=0.25cm, height= 0.25cm]{gray_images-rounded.png}  &   & & WWW 2020 \\  
  \hline
\bluer{G2ANet} \cite{liu2020multi} & AG &
		\hspace{0.20cm} \includegraphics[width=0.25cm, height= 0.25cm]{gray_images-rounded.png} 
		&  \hspace{0.50cm} \includegraphics[width=0.25cm, height= 0.25cm]{gray_images-rounded.png}
  & \hspace{0.50cm} \includegraphics[width=0.25cm, height= 0.25cm]{gray_images-rounded.png}  &  \hspace{0.50cm} \includegraphics[width=0.25cm, height= 0.25cm]{gray_images-rounded.png} & & AAAI 2020 \\  
  \hline
  	\bluer{DCG} \cite{bohmer2020deep} & AG &
		\hspace{0.20cm} \includegraphics[width=0.25cm, height= 0.25cm]{gray_images-rounded.png} 
		 & \hspace{0.50cm} \includegraphics[width=0.25cm, height= 0.25cm]{gray_images-rounded.png} 
  &  \hspace{0.50cm} \includegraphics[width=0.25cm, height= 0.25cm]{gray_images-rounded.png}  &  \hspace{0.50cm} \includegraphics[width=0.25cm, height= 0.25cm]{gray_images-rounded.png}   &\hspace{0.50cm} \href{https://github.com/wendelinboehmer/dcg}{\includegraphics[width=0.25cm, height= 0.25cm]{gray_images-rounded.png}} & ICML 2020 \\  
  \hline
  	\bluer{GraphComm} \cite{shen2021graphcomm} & AG &
		\hspace{0.20cm} \includegraphics[width=0.25cm, height= 0.25cm]{gray_images-rounded.png} 
	  & 	\hspace{0.50cm} \includegraphics[width=0.25cm, height= 0.25cm]{gray_images-rounded.png}
  &   & \hspace{0.50cm} \includegraphics[width=0.25cm, height= 0.25cm]{gray_images-rounded.png}  & & ICASSP 2021 \\  
  \hline
  	\bluer{SRI-AC} \cite{zhang2021structural} & AG &
		\hspace{0.20cm} \includegraphics[width=0.25cm, height= 0.25cm]{gray_images-rounded.png} 
		& \hspace{0.5cm} \includegraphics[width=0.25cm, height= 0.25cm]{gray_images-rounded.png} 
  &   & \hspace{0.5cm} \includegraphics[width=0.25cm, height= 0.25cm]{gray_images-rounded.png}   &  & Elsevier 2021 \\  
  \hline
  	\bluer{CSGA} \cite{yun2021multi} & AG &
		\hspace{0.20cm} \includegraphics[width=0.25cm, height= 0.25cm]{gray_images-rounded.png} 
		& \hspace{0.5cm} \includegraphics[width=0.25cm, height= 0.25cm]{gray_images-rounded.png}  & \hspace{0.50cm} \includegraphics[width=0.25cm, height= 0.25cm]{gray_images-rounded.png} 
  &  	\hspace{0.50cm} \includegraphics[width=0.25cm, height= 0.25cm]{gray_images-rounded.png}  &  & IEEE SMC-2021 \\  
  \hline
    	\bluer{NerveNet} \cite{mtdrl_02} & AG &
		\hspace{0.20cm} \includegraphics[width=0.25cm, height= 0.25cm]{gray_images-rounded.png} 
		& \hspace{0.5cm} \includegraphics[width=0.25cm, height= 0.25cm]{gray_images-rounded.png}  & \hspace{0.50cm} \includegraphics[width=0.25cm, height= 0.25cm]{gray_images-rounded.png} 
  & \hspace{0.50cm} \includegraphics[width=0.25cm, height= 0.25cm]{gray_images-rounded.png}   & \hspace{0.50cm} \href{https://github.com/WilsonWangTHU/NerveNet}{\includegraphics[width=0.25cm, height= 0.25cm]{gray_images-rounded.png}}  & ICLR-2018 \\  
  \hline
    	\bluer{SMP} \cite{mtdrl_03} & AG &
		& \hspace{0.5cm} \includegraphics[width=0.25cm, height= 0.25cm]{gray_images-rounded.png}  & \hspace{0.5cm} \includegraphics[width=0.25cm, height= 0.25cm]{gray_images-rounded.png}
  &  \hspace{0.5cm} \includegraphics[width=0.25cm, height= 0.25cm]{gray_images-rounded.png}   & \hspace{0.50cm} \href{https://github.com/huangwl18/modular-rl}{\includegraphics[width=0.25cm, height= 0.25cm]{gray_images-rounded.png}}   & ICML-2020 \\  
  \hline
      	\bluer{SymNet} \cite{rdrl_04} & AG &
\hspace{0.2cm} \includegraphics[width=0.25cm, height= 0.25cm]{gray_images-rounded.png} 
		& \hspace{0.5cm} \includegraphics[width=0.25cm, height= 0.25cm]{gray_images-rounded.png}  & \hspace{0.5cm} \includegraphics[width=0.25cm, height= 0.25cm]{gray_images-rounded.png}
  &  \hspace{0.5cm} \includegraphics[width=0.25cm, height= 0.25cm]{gray_images-rounded.png}   & \hspace{0.50cm} \href{https://github.com/dair-iitd/symnet}{\includegraphics[width=0.25cm, height= 0.25cm]{gray_images-rounded.png}}   & ICML-2020 \\  
  \hline
      	\bluer{SR-DRL} \cite{rdrl_05} & AG & \hspace{0.2cm} \includegraphics[width=0.25cm, height =  0.25cm]{gray_images-rounded.png} 
		& \hspace{0.5cm} \includegraphics[width=0.25cm, height= 0.25cm]{gray_images-rounded.png}  & \hspace{0.5cm} \includegraphics[width=0.25cm, height= 0.25cm]{gray_images-rounded.png}
  & & \hspace{0.50cm} \href{https://github.com/jaromiru/sr-drl}{\includegraphics[width=0.25cm, height= 0.25cm]{gray_images-rounded.png}}   & arxiv 2021 \\ \hline 
  \multicolumn{8}{|c|}{\bf Applications} \\ \hline
  	\bluer{GCOMB} \cite{manchanda2020gcomb} & PC &
		\hspace{0.20cm} \includegraphics[width=0.25cm, height= 0.25cm]{gray_images-rounded.png} 
		& \hspace{0.5cm} \includegraphics[width=0.25cm, height= 0.25cm]{gray_images-rounded.png}  & \hspace{0.5cm} \includegraphics[width=0.25cm, height= 0.25cm]{gray_images-rounded.png} 
  &   &  \hspace{0.50cm} \href{https://github.com/idea-iitd/GCOMB}{\includegraphics[width=0.25cm, height= 0.25cm]{gray_images-rounded.png}}  & NeurIPS-2020 \\  
  \hline
    	Khalil et al. \cite{khalil2017learning} & PC &
		& \hspace{0.5cm} \includegraphics[width=0.25cm, height= 0.25cm]{gray_images-rounded.png}  & \hspace{0.50cm} \includegraphics[width=0.25cm, height= 0.25cm]{gray_images-rounded.png}
  &  & \hspace{0.5cm} \href{https://github.com/Hanjun-Dai/graph_comb_opt}{\includegraphics[width=0.25cm, height= 0.25cm]{gray_images-rounded.png}} & NeurIPS-2017 \\  
  \hline
    \bluer{GraMeR} \cite{Munikoti2022GraMeR} & PC &
		& \hspace{0.5cm} \includegraphics[width=0.25cm, height= 0.25cm]{gray_images-rounded.png}  & \hspace{0.5cm} \includegraphics[width=0.25cm, height= 0.25cm]{gray_images-rounded.png}
  & \  & \hspace{0.5cm} \href{https://github.com/saimunikoti/InfluenceMaximization-Deep-QLearning}{\includegraphics[width=0.25cm, height= 0.25cm]{gray_images-rounded.png}}  & arxiv 2022 \\  
  \hline
      	\bluer{RLGN} \cite{meirom2021controlling} & PC &
		\hspace{0.20cm} \includegraphics[width=0.25cm, height= 0.25cm]{gray_images-rounded.png} 
		& \hspace{0.5cm} \includegraphics[width=0.25cm, height= 0.25cm]{gray_images-rounded.png}  & \hspace{0.5cm} \includegraphics[width=0.25cm, height= 0.25cm]{gray_images-rounded.png} 
  &   &  & ICML-2021 \\  
  \hline
       \bluer{GCN-RL} \cite{wang2020gcn} & PC &
		& \hspace{0.5cm} \includegraphics[width=0.25cm, height= 0.25cm]{gray_images-rounded.png}  & \hspace{0.5cm} \includegraphics[width=0.25cm, height= 0.25cm]{gray_images-rounded.png}
  &   &  & DAC-2020 \\  
  \hline
        \bluer{Rnet-DQN} \cite{darvariu2021goal} & PC &
		& \hspace{0.5cm} \includegraphics[width=0.25cm, height= 0.25cm]{gray_images-rounded.png}  & \hspace{0.5cm} \includegraphics[width=0.25cm, height= 0.25cm]{gray_images-rounded.png}
  & \  & \hspace{0.5cm} \href{https://github.com/VictorDarvariu/graph-construction-rl}{\includegraphics[width=0.25cm, height= 0.25cm]{gray_images-rounded.png}}  & PRSA-2021 \\  
  \hline
    Drori et al. \cite{drori2020learning} & PT &
		& \hspace{0.5cm} \includegraphics[width=0.25cm, height= 0.25cm]{gray_images-rounded.png}  & \hspace{0.5cm} \includegraphics[width=0.25cm, height= 0.25cm]{gray_images-rounded.png}
  &   &  & ICMLA-2020 \\  
  \hline
       \bluer{L2I}\cite{lu2019learning} & PT &
		& \hspace{0.5cm} \includegraphics[width=0.25cm, height= 0.25cm]{gray_images-rounded.png}  & 
  &   & \hspace{0.5cm} \href{https://github.com/rlopt/l2i}{\includegraphics[width=0.25cm, height= 0.25cm]{gray_images-rounded.png}}  & ICLR-2020 \\  
  \hline
        Hu et al. \cite{hu2020reinforcement} & PT &
		& \hspace{0.5cm} \includegraphics[width=0.25cm, height= 0.25cm]{gray_images-rounded.png}  & \hspace{0.5cm} \includegraphics[width=0.25cm, height= 0.25cm]{gray_images-rounded.png}
  &   &  & Elsevier-2020 \\  
  \hline
           \bluer{IG-RL} \cite{devailly2021ig} & PT &
		\hspace{0.20cm} \includegraphics[width=0.25cm, height= 0.25cm]{gray_images-rounded.png} 
		& \hspace{0.5cm} \includegraphics[width=0.25cm, height= 0.25cm]{gray_images-rounded.png}  & \hspace{0.5cm} \includegraphics[width=0.25cm, height= 0.25cm]{gray_images-rounded.png}
  &   & \hspace{0.5cm} \href{https://github.com/FXDevailly/IG-RL}{\includegraphics[width=0.25cm, height= 0.25cm]{gray_images-rounded.png}}   & IEEE-2021 \\ 
  \hline
         Shang et al. \cite{shang2022new} & PT &
		\hspace{0.20cm} \includegraphics[width=0.25cm, height= 0.25cm]{gray_images-rounded.png} 
		& \hspace{0.5cm} \includegraphics[width=0.25cm, height= 0.25cm]{gray_images-rounded.png}  & \hspace{0.5cm} \includegraphics[width=0.25cm, height= 0.25cm]{gray_images-rounded.png}
  &   &  & Elsevier-2022 \\  
  \hline
            Lima et al. \cite{lima2020ieeeresource} & PR &
		\hspace{0.20cm} \includegraphics[width=0.25cm, height= 0.25cm]{gray_images-rounded.png} 
		& \hspace{0.5cm} \includegraphics[width=0.25cm, height= 0.25cm]{gray_images-rounded.png} & \hspace{0.5cm} \includegraphics[width=0.25cm, height= 0.25cm]{gray_images-rounded.png}
  &   &   & IFAC-2020 \\  
  \hline
          Wang et al. \cite{wang2020multi} & PR &
		\hspace{0.20cm} \includegraphics[width=0.25cm, height= 0.25cm]{gray_images-rounded.png} 
		& \hspace{0.5cm} \includegraphics[width=0.25cm, height= 0.25cm]{gray_images-rounded.png}  & \hspace{0.5cm} \includegraphics[width=0.25cm, height= 0.25cm]{gray_images-rounded.png}
  &   &   &  IFAC-2020 \\  
  \hline
            \bluer{GNN-MARL} \cite{huang2021integrated} & PR &
		\hspace{0.20cm} \includegraphics[width=0.25cm, height= 0.25cm]{gray_images-rounded.png} 
		&   & 
  &   &  & CIRP-2021 \\  
  \hline
             Park et al. \cite{park2021learning} & PR &
		\hspace{0.20cm} \includegraphics[width=0.25cm, height= 0.25cm]{gray_images-rounded.png} 
		& \hspace{0.5cm} \includegraphics[width=0.25cm, height= 0.25cm]{gray_images-rounded.png}  & \hspace{0.5cm} \includegraphics[width=0.25cm, height= 0.25cm]{gray_images-rounded.png}
  &   &   & T\&F-2021 \\  
  \hline
          \bluer{GCQ} \cite{chen2021graph} & PR &
		\hspace{0.20cm} \includegraphics[width=0.25cm, height= 0.25cm]{gray_images-rounded.png} 
		&  & 
  & \hspace{0.5cm} \includegraphics[width=0.25cm, height= 0.25cm]{gray_images-rounded.png}  &   & Wiley-2021 \\  
  \hline
       \bluer{GRL} \cite{kgc_reinLrn_03} & PK &
		\hspace{0.20cm} \includegraphics[width=0.25cm, height= 0.25cm]{gray_images-rounded.png} 
		& \hspace{0.5cm} \includegraphics[width=0.25cm, height= 0.25cm]{gray_images-rounded.png}  & \hspace{0.5cm} \includegraphics[width=0.25cm, height= 0.25cm]{gray_images-rounded.png} 
  &   &   & Elsevier-2020 \\  
  \hline
   \bluer{3D-MolGNNRL} \cite{mcnaughton2022novo} & PL &
		\hspace{0.20cm} \includegraphics[width=0.25cm, height= 0.25cm]{gray_images-rounded.png} 
		& \hspace{0.5cm} \includegraphics[width=0.25cm, height= 0.25cm]{gray_images-rounded.png}  & \hspace{0.5cm} \includegraphics[width=0.25cm, height= 0.25cm]{gray_images-rounded.png}
  &   &  & ICLR-2022 \\  
  \hline
     \bluer{GTPN} \cite{do2019graph} & PL &
		& \hspace{0.5cm} \includegraphics[width=0.25cm, height= 0.25cm]{gray_images-rounded.png}  & \includegraphics[width=0.25cm, height= 0.25cm]{gray_images-rounded.png}
  &  &   & KDD-2019 \\  
  \hline
\end{tabular}
\end{table*}

\begin{table*}
	\scriptsize
	\centering
	\caption{DRL and GNN components of the surveyed papers}
	\label{tab:qualitativesummary}
	\resizebox{2.03\columnwidth}{!}{
	\begin{tabular}{|p{1.0cm}|p{0.5cm}|p{2.0cm}|p{2.5cm}|p{2.0cm}|p{1.2cm}|p{1.3cm}|p{3.5cm}|}
		\hline
		 \textbf{Ref.} & \textbf{Categ.} &\textbf{State} & \textbf{Action} & \textbf{Reward}  & \textbf{DRL}  & \textbf{GNN} & \textbf{Remarks}  \\
\hline
\multicolumn{8}{|c|}{\bf Algorithms} \\ \hline
	     \bluer{AGNN} \cite{gao2020graph} & AD & Activation function in GNN layers & Changes in activations, aggregation, hidden units, \# heads & Validation accuracy & PPO & GCN, GAT, LGCN & LSTM encodes graph architecture \& DRL for optimizing accuracy \\
\hline
	      \bluer{RG-Explainer} \cite{shan2021reinforcement} & AD & Node feature and current subgraph & validation performance + controller entropy & Prediction loss &  GraphSAGE & GAT & Identify most influential subgraph to interpret node prediction \\
\hline
	     \cite{sun2020non} & AD & Intermediate poisoned graph & Add adversarial edge \& change labels of injected nodes & Node classification accuracy & DQN &  GCN & novel non-target specific node injection poisoning attack on graphs \\ 
\hline
	      \bluer{CSGA} \cite{yun2021multi} & AG & characteristics of agents; partial information of other agents and enemies & cooperative actions of individual agents & joint-action value function & DRQN &  GAT & Novel MADRL algorithm to control multiple agents in RTS games. \\
\hline
	     \bluer{SMP} \cite{mtdrl_03} & AG & positions, velocity, rotations & position with lower and upper bound & actuator response & DDPG & Message passing & shared modular policy that is generalize to variants ( several planar agents with different skeletal structures) not seen during training \\
\hline
	      \bluer{SR-DRL}\cite{rdrl_05} & AG & objects and their features, relations, global context & object labels, set of parameters, preconditions & goal specifications & A2C &  GAT & Generic framework for solving relational domains \\
\hline
        \multicolumn{8}{|c|}{\bf Applications} \\ 
\hline
	      \bluer{GCOMB}\cite{manchanda2020gcomb} & PC & candidate nodes in solution set & Adding a node to solution set & marginal gain of action function & DQN & GCN & Scalable constrained learning of combinatorial algorithms \\
\hline
	     \cite{khalil2017learning} & PC & nodes in solution set & Adding a node to solution set & marginal gain in cost function & DQN & structure2vec &  First learning framework for graph combinatorial \\
\hline
	      \bluer{GraMeR}\bluer{\cite{Munikoti2022GraMeR}} & PC & nodes in solution set & select the node from candidate set & marginal gain in cost function & Double DQN & GraphSage &  GNN predictor to reduce the search space of the DRL agent \\
\hline
	      \bluer{RNet-DQN} \cite{darvariu2021goal} & PC & Graph and edge stub(node) & Adding a node to solution set & gain in graph robustness score & DQN & structure2vec &  Learning graph construction using DRL and GNN \\
\hline
	     \cite{deudon2018learning} & PT & Partial set of visited cities & next city to visit & Tour length & REINFORCE & GAT & Learning clever heuristics  \\
\hline
	     \bluer{IG-RL} \cite{devailly2021ig} & PT & Current Connectivity and demand & Switch to next phase or continue with current & number of vehicles stopped on a lane & Double \newline Duelling DQN & GCN & Inductive learning applicable to various road networks \& traffic distribution \\
\hline
	     \cite{shang2022new} & PT & GNN parameters & Modify GNN parameters & Target function error &  DQN & GCN \& GAT & Ensemble model weighing GCN and GAT predictions \\
\hline
	     \cite{wang2020multi} & PR & Agents local observations & Unit movement in X \& Y directions & Distance from target point &  REINFORCE & REGNN & Leverages graph attention \& LSTM for cooperative agent behavior  \\
\hline
	     \bluer{GNN-MARL} \cite{huang2021integrated} & PR & node (machine) latent features  & machine control settings & difference between step wise yield \& defect &  C-COMA & GCN & Distributed adaptive control for multistage systems \\
\hline
	     \cite{park2021learning} & PR & Snapshot of job shop at specific transition & Loading/skipping scheduling & negative of number of waiting jobs &  PPO & GCN & Generic framework for Job shop scheduling incorporating spatial and temporal structure \\
\hline
	     \cite{wang2021reinforcement} & PK & Combination of entity and relation space & Selecting neighboring relational path to another entity & Closeness of new state to target entity & DDPG & GCN & Leverages GAN and LSTM to capture graph dynamics for knowledge graph completion \\
\hline
	      \bluer{GRL} \cite{kgc_reinLrn_03} & PK & Combination of entity and relation space & Selecting neighboring relational path to another entity & Closeness of new state to target entity & A2C & GCN & knowledge graph completion with domain specific rules\\
\hline
	      \bluer{3D-MolGNNRL} \cite{mcnaughton2022novo} & PL & Intermediate molecule & Add new atom to partial molecule & binding probability and affinity & A2C & GAT & include 3D structure of both protein target and generated compounds for drug design.\\
\hline
	     \bluer{GTPN} \cite{do2019graph} & PL & Intermediate compound & add new bond between an atom pair & +/- 1 whether generated product matches groundtruth & A2C & GAT & Uses GNN to represent reactant/reagent molecules, and DRL to find an optimal sequence of bond changes for product transformation\\
\hline
\end{tabular}
}
\end{table*}

\section{Discussion and lessons learned}

\label{sec:lessons}
Supported by an extensive review, we observe that the use of GNNs in a DRL framework is becoming increasingly popular from an algorithmic development perspective and applications of machine learning to complex problems. In this section, we present our perspectives in terms of applicability and advantages of fusing these learning frameworks. 

\subsection{Advantages of fusing DRL and GNN}

As discussed before, GNN and DRL are fused in two different fronts, i.e., algorithmic enhancement where methodologies are enhancing each other and application where algorithms are supporting each other. This fusion has several advantages that can be summarized as follows: \\
(\textbf{1}) On moving from single agent to multi-agent or from single task to multi-task scenarios in DRL, the complexity of problem drastically increases. Therefore, various new approaches are continuously being proposed to improve the model performance. However, there is always a scope to incorporate auxiliary information for further improvement. Since MADRL/MTDRL involves multiple agents, incorporating the relational information among these agents in the core model with a GNN architecture can improve its performance. 
\textit{
Since GNNs are inherently designed to capture topological/attributed relationships, they are powerful models that allow capturing the multi-agent and multi-task relationships relative to other models};\\
(\textbf{2}) GNN, like other DNN models require further improvements in terms of automatic setting generation, improving model explainability and enhancing robustness against adversarial attacks. These tasks can easily be handled via DRL due to inherent sequential nature. \textit{ DRL is well-suited compared to traditional optimization based approaches for these tasks since it offers a computationally lightweight framework to tackle large problem spaces in a scalable and generic way};\\
(\textbf{3}) The performance of DRL in applications involving graph-structured environment such as knowledge graphs and transportation networks depend on encoder to a large extent. Therefore, GNNs are used to represent trajectory information in such environments and also act as a function approximator. \textit{ GNNs are very effective in representing/encoding graphs compared to other techniques such as graph signal processing or spectral graph theoretic approaches. Further, they are flexible and generic enough to work for different graph families and sizes}.   

\subsection{Problem-specific applicability of DRL and GNN methods}

A fusion of GNN and DRL has found itself a set of niche problems that span diverse applications, while sharing common features. These common features are: (\textbf{1}) \bluer{\textit{Sequential decision making}} setting of the problem, wherein learning occurs via interactions with the environment in a closed loop manner; (\textbf{2}) The learning agent \bluer{\textit{exploits}} its acquired knowledge at any time, while also striking a balance between \bluer{\textit{exploring}} multiple options for a potentially better solution; (\textbf{3}) The learning is aimed at achieving \bluer{\textit{long term goals}} and avoid making myopic decisions; (\textbf{4}) The underlying system is most efficiently represented as a \bluer{\textit{graph}}, thereby making GNNs the natural choice for representing such systems. A widely studied example of such a problem is the traveling salesperson problem, where the process of finding the optimal route is a sequential process of identifying nodes that lead to minimum total distance travelled. Furthermore, the underlying problem possesses a graph structure with nodes being destinations and links representing connection between them.


The majority of applications in literature involve \bluer{\textit{static systems}}, so that a single GNN module can serve as both a function approximator and an encoder for the environment. However, depending on the nature of the problem, an appropriate GNN algorithm must be chosen for the best performance. Environments involving large graphs should rely on GraphSAGE \cite{hamilton2017inductive} rather than GCN \cite{kipf2016semi}, as GraphSAGE is a sub-graph-based inductive learning approach that is scalable to larger networks. Similarly, applications where the position of a node with respect to the entire graph is vital, position-aware GNNs (PGNN) \cite{you2019position} are preferred. PGNN  explicitly make use of anchor nodes along with neighboring sub-graph to improve the effectiveness of node embeddings. Furthermore, the expressive power of most GNNs is upper-bounded by the 1-Weisfeiler-Lehman (1-WL) graph isomorphism test, i.e., they cannot differentiate between different d-regular graphs. Therefore, it is recommended to explore identity-aware GNN \cite{you2021identity} for complex graph-structured environments, which inductively considers nodes’ identities during message passing.   

Along with static graphs, there are certain applications that involve \bluer{\textit{dynamic graph-structured environments}}. For instance, in a connected autonomous vehicle network, the number of vehicles dynamically changes. Under this scenario, an appropriate strategy would be to use LSTMs fused with GNNs for capturing graph evolution as well as DRL trajectories. At any instant, the spatial information of the environment can be gathered via GNN and fed to an LSTM cell state for learning long-range spatio-temporal dependency. Furthermore, separate GNNs can be used to encode topological changes and long-range dependencies individually. \bluer{Dynamic graphs can also be handled via novel graph neural odes formulation (GDE), where input-output relationship is determined by a continuum
of GNN layers, blending discrete topological structures and differential equations \cite{poli2019graph,zhuang2019ordinary}.  The structure–dependent vector field learned by GDEs offers a data–driven approach to the modeling of dynamical networked
systems, particularly when the governing equations
are highly nonlinear and therefore challenging to approach with analytical methods \cite{furieri2022distributed}. Autoregressive
GDE can adapt the prediction horizon by adjusting the integration interval of the ODE, allowing
the model to track the evolution of the underlying system from irregular observations \cite{poli2019graph}} In addition to the type of environment, the problem of interest can have a single learning agent or multiple agents. In a multi-agent application without any interaction between agents, traditional MADRL algorithms are most suited. However, in certain scenarios, the agents might interact with each other in search of a better solution. These interactions can further be predefined or might appear as an agent interacts with the environment. GNN models can be used to capture such relationships and provide auxiliary information to the agent in order to further improve model performance. This also applies to multi-task situations where different tasks are correlated or structurally related.  

GNNs encompasses various tuning parameters in their architecture. Thus, \bluer{\textit{neural architecture search}} in them are very effective. DRL algorithms such as DQN are an appropriate choice for searching since the search methodology is generic and applicable across different architectures. In fact, any applications involving search operations in GNN such as adversarial attack can be tackled very effectively with DRL. The use of multimodal data has led to heterogeneous graph-structured data in various applications, including knowledge graphs and recommender systems. Conventional GNNs are not designed to handle heterogeneity. Therefore, it is recommended to employ customized GNNs like relational GCN \cite{schlichtkrull2018modeling}, heterogeneous GAT \cite{wang2019heterogeneous} and HetSANN \cite{hong2020attention} for encoding. Fundamentally, all these works demand separate aggregation and combination functions (model parameters) for each node/link type, i.e., node attributes or link relationships, so that a powerful node representation can be learned.

\section{Challenges and Future Research opportunities}

\label{sec:challenges}
\begin{figure}[h!]
	\centering
	\includegraphics[width=\columnwidth]{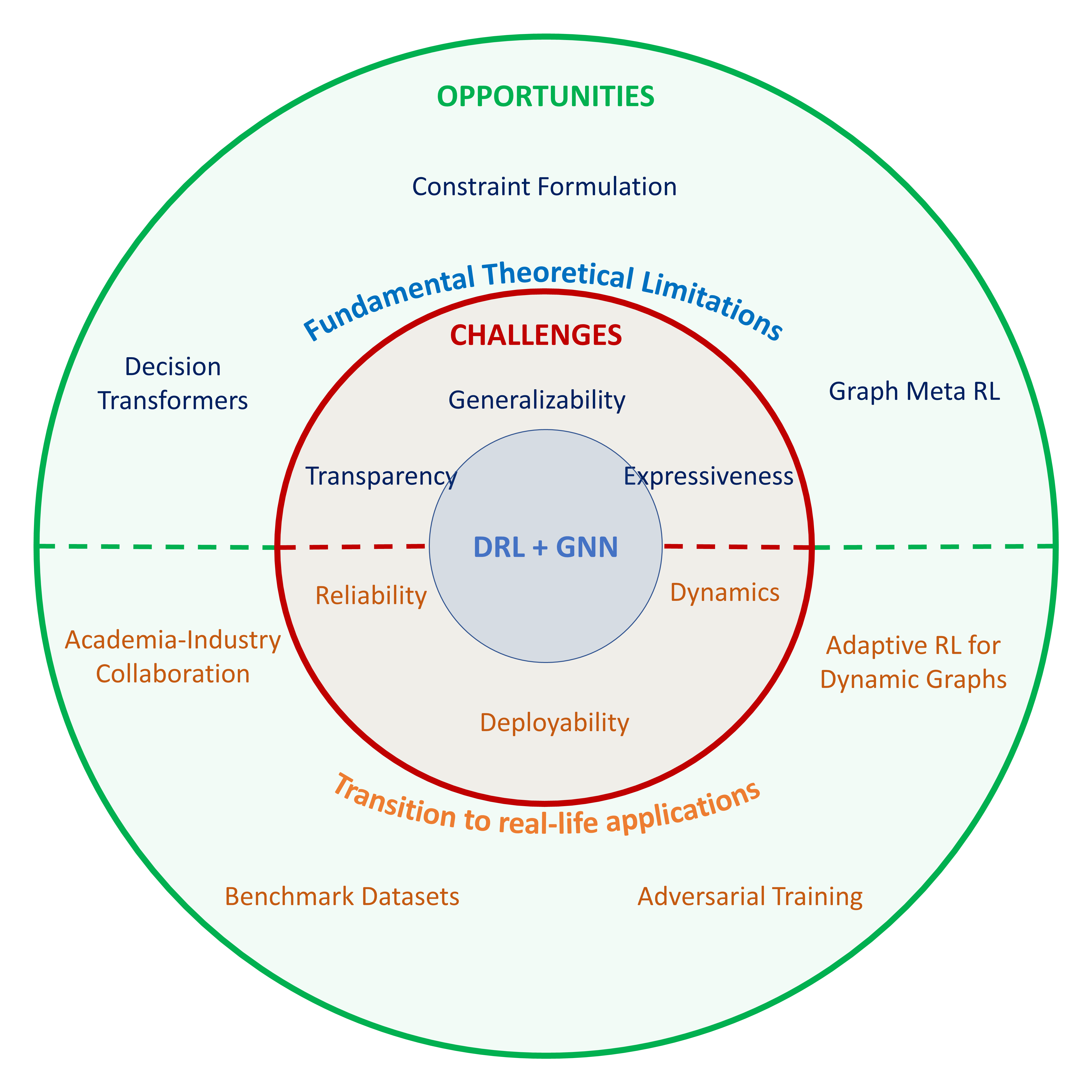}
	\caption{\bluer{Challenges and Future Research Opportunities in hybrid DRL-GNN paradigm.}}
	\label{fig:oprtChlng}
\end{figure}

The articles surveyed in this work reveal the wide applicability and importance of fusing GNN and DRL (summarized in \bluer{(Section \RomanNumeralCaps{4})}. This section identifies the challenges that lie ahead for widespread adoption, and suggests future directions to unlock the full potential of a combined GNN-DRL framework.\bluer{These challenges and opportunities are shown in Figure \ref{fig:oprtChlng}. We identify two broad classes of challenges - those associated with fundamental theoretical limitations of the existing approaches such as generalizability, explainability, expressivity (interpretability), and the ones concerned with the transfer of techniques to real life applications, e.g. standard benchmark design, sensitivty analyses, uncertainty and reliabilty quantification, etc.}

\bluer{\subsection{Improving generalizability with state-of-the-art models and learning frameworks}}
\bluer{Deep neural networks are notoriously challenging in terms of generalizing to different problems. This problem predominantly stems from the large size, and hence model capacity of modern deep neural networks. A combined GNN-DRL approach can fall into the pitfall of poor generalzability by not being able to generalize to slightly different environment settings. For instance, changing the layout of environment (e.g. dynamic link connections in social network while solving combinatorial problem such as influence maximization) for a model that learns to navigate through the environment can result in the model being stuck in a segment of the network, or to take large amount of time to navigate the network. This can be caused either due to the GNN, or the DRL not being able to generalize to the environment.}

\bluer{One possible approach to address this challenge is to use {\em graph meta reinforcement learning framework} where agents can quickly adapt to new tasks or environment with fewer samples~\cite{Munikoti2022GraMeR}.}
Specifically, this can be achieved by providing context variables that vary with applications\bluer{/environments}. For example, in case of CO, variations could be the problems that are smaller instances of the same problem, problem instances with different distributions, or even the ones from the other types of CO problems. Although certain generalization efforts can be seen recently, there is more to be done. 

Another way to enhance generalizability of DRL algorithms is by exposing agents to several graph environments that can be created with training graphs via graph augmentation techniques. One potential idea would be in leveraging GAN for graph augmentation by generating synthetic examples via perturbing input graphs in terms of adding/removing nodes and links or modifying node/link attributes \bluer{\cite{fan2019dismantle,fan2020novel}}. This enables DRL agent to adapt by learning invariant features across varied and noisy environments. We believe that although this task is challenging, it is extremely important and a promising direction for research in hybird DRL-GNN works.  
\bluer{Moreover, recently, researchers brought the idea of Decision Transformers (similar to large pre-trained language models such as BERT, GPT, etc.) to DRL algorithms \cite{zheng2022online}. This involves pre-training on large set of environments and then fine-tuning on the specific application. It will be interesting to explore these pre-training and fine-tuning tasks for graph-structured environments. }

\bluer{\subsection{Increasing transparency by incorporating XAI and representation learning}}
A substantial amount of explainable Artificial Intelligence (XAI) literature is emerging on feature relevance techniques to explain the predictions of a deep neural network (DNN) or explaining models that consume image source data. However, it is unclear as to how  XAI techniques can aid in understanding models beyond classification tasks, such as DRL. An improved interpretability and explainability of DRL (XRL) models could help shed light on the underlying mechanisms in situations where it is essential to justify and explain the agent's behavior, which is still contemplated as a black box. The recent efforts on XRL are problem specific and cannot be generalized to real-world RL tasks \cite{expl_02}. \bluer{Explaining predictions of GNN and  explaining reasoning behind the DRL actions are important and active area of research.}

The concepts based on representation learning, like hierarchical RL, self-attention and Hindsight Experience Replay have been highlighted as a few encouraging approaches to improve explainability in DRL models \cite{expl_02}, and can be considered as future lines of research. 
Furthermore, DRL is also employed for explaining node/link predictions in GNN by identifying the most influential sub-graph \cite{shan2021reinforcement}.  
The use of GNN to improve the prediction of DRL can also be investigated as a part of future research in this space.    

\bluer{\subsection{Incorporating network constraints with novel formulations}}

Optimization problems for real-life applications are mostly bounded by a variety of constraints in terms of finance, time, resources, etc. Most of the existing DRL work deal with constraints through penalties in rewards, which is appropriate if the constraints are soft, i.e., they can be violated at some cost. However, hard constraints must be strictly met and imposing a penalty cannot eliminate them, and thus is not a perfect approach. An alternative way to deal with hard constraints is by masking the constraints while designing the training environment, to keep the exploration space away from constraint violation, as considered in autonomous driving \cite{nazari2018reinforcement, liu2021policy}. These constraints become further complex when the underlying environment is graph-structured, such as a transportation network. \bluer{ A few attempts have been made in this direction with remarkable performance such as reward shaping via hierarchical framework \cite{cappart2021combining} and deferred markov decision process \cite{ahn2020learning}. However, there is still a lot of scope to improve.} Therefore, further research is needed to explore the rich literature of constrained dynamical systems and other strategies for dealing with hard constraints.

\bluer{\subsection{Learning with dynamic/heterogeneous environment}}
A majority of the existing GNN models perform prediction and inferencing over homogeneous graphs. However, a large number of real-world applications, like critical infrastructure networks, recommendation systems, and social networks, involve learning on heterogeneous graphs. Heterogeneous graph-structured data can represent numerous types of entities (nodes) and relations (edges) within a common framework. It is difficult to handle such diverse graphs using existing GNN models. Consequently, developing new models and algorithms that are capable of learning with heterogeneous graphs would be highly beneficial in real-world systems like cybersecurity \cite{dynaHet_01}, \cite{dynaHet_02}, text analysis \cite{dynaHet_03}, \cite{dynaHet_04} and recommendation engines \cite{dynaHet_05}, \cite{dynaHet_06}. Integrating the use of DRL techniques to achieve this goal can be considered as one of the possible future directions of research. Furthermore, the existing GNN methodologies assume graph-structured data to be static, i.e., the possibility of addition and/or removal of nodes/edges is disregarded. However, many practical applications like social networks encompass dynamic spatial relations that continuously evolve over time. Although spatial-temporal GNNs (STGNNs) possess the ability to partially handle dynamic graphs \cite{dynaHet_07}, further work is required to integrate a thorough understanding about executing downstream tasks like node classification, link prediction, community detection, and graph classification in dynamic graphs.

\bluer{\subsection{Enabling a seamless transition to real-world networks}}

Most of the prevailing GNN-DRL methods are developed based on synthetic graph-structured datasets and simulated platforms. \bluer{However,} real-life scenarios \bluer{can be } far more intricate compared to simulated platforms. 
Thus, \bluer{the resulting models} need rigorous validation before being deployed confidently in real life applications. This validation is specifically more crucial for applications like connected autonomous vehicle, manufacturing process, etc., where safety is of utmost importance. There are some attempts to transfer the trained DRL agent from a simulator to an actual test bed \cite{trnfDRL_simReal_01} in general DRL, but there is a significant gap for graph-structured environments \bluer{in general}. Therefore, seamless transition from a simulated setting to real-world scenarios in a cautious, protected and productive approach is an important future research direction. \bluer{ In this regard, more collaborative efforts from the industry and academia through partnership projects is required to understand the limitations of models in real-life applications and devise methods to improve the model's performance. It is important to design standardised problem settings and benchmark datasets that can be used to evaluate and compare different methodologies for specific applications, enabling the faster development of the state of the art solutions.}

\bluer{\subsection{Assessing transferrability with uncertainty and reliability quantification}}
From the perspective of practitioners, it is critical that \bluer{any solution be robust to slight variations in its inputs. Specifically in the case of hybrid DRL-GNN approaches, it is important that the }solutions be robust to changes in \bluer{the } graph and environment. \bluer{A sensitivity analysis of the model's performance to data and environmental parameters can provide important insights into the stability of a model, and help identify }
suitable modifications in terms of environment design, model specification, training process, and data fidelity among others \bluer{to improve the real-life applicability of the model. In addition, quantifying the uncertainty of the model will allow an end-user to calibrate and evaluate the confidence in the model's predictions. This can be achieved with Bayesian approaches such as Bayes by backpropagation, assumed density filtering, Monte-Carlo dropout, or by ensemble methods \cite{munikoti2022general} . While several of these studies have been performed for DNNs, a lot of progress is still required in DRL-GNN approaches to assess the applicability to real-life scenarios. }
\bluer{Finally, generative models such as GANs can be used to generate different scenarios and potentially improve the model's performance.}

\bluer{\subsection{Computational Requirements}
Schwartz et al., observe that from AlexNet to AlphaZero, the computational requirements have grown more than $300,000\times$ in six years \cite{GreenAI}. For example, a single game of AlphaGo required 1,920 CPUs and 280 GPUs with an estimated cost of $\$35,000,000$ \cite{GreenAI}. Given the large parameter space, as well as the lack of formal complexity analysis in existing literature, it remains challenging to accurately assess the computational requirements of combined DRL and GNN models. 
We posit it that the complexity of the combined models will be additive in nature, rather than multiplicative, and therefore, will be bounded by the larger of the two complexities of GNN or DRL models for a given problem setting. Recently, the concept of Foundation model has become popular for NLP, where single large pretrained model can be efficiently fine tuned for several downstream tasks \cite{bommasani2021opportunities}. We believe that a single Foundation model can also be developed for DRL-GNN paradigm with high few/zero-shot learning capability that can reduce the computational burden imposed by training several problem-specific decision models. }

\section{Summary and Conclusions}

\label{sec:conclusions}
This paper presents a systematic survey of literature focusing on works fusing DRL and GNN approaches. Although several reviews related to DRL have been published in recent years, most of these studies are limited to a particular application domain. This study, for the first time, presents a methodical review spanning diverse range of application domains. We have reviewed papers from the perspectives of both fundamental algorithmic enhancements as well as application-specific developments. From an algorithmic perspective, either a GNN is exploited to strengthen the formulation and improve performance of DRL, or DRL is employed to expand the applicability of GNN. Recent works blending the usage of both DRL and GNN across multiple applications (broadly classified into combinatorial optimization, transportation, manufacturing and control, knowledge graphs and life sciences) have been thoroughly investigated and discussed. We also highlight the key advantages of fusing DRL and GNN methodologies and outline the applicability of each of those components. Furthermore, we identify the challenges involved in effective integration of DRL and GNN, and propose some potential future research directions in this area.

\mbox{}
\nomenclature[01]{$X$}{State of a MDP}
\nomenclature[02]{$A$}{Action space of MDP}
\nomenclature[03]{$R$}{Reward in a MDP}
\nomenclature[04]{$S$}{Solution set}
\nomenclature[04]{$\Pi$}{Policy}
\nomenclature[A, 01]{MDP}{Markov decision process}
\nomenclature[A, 02]{DRL}{Deep reinforcement learning}
\nomenclature[A, 03]{GNN}{Graph Neural Network}
\nomenclature[A, 04]{DNN}{Deep Neural Network}
\nomenclature[A, 05]{RL}{Reinforcement learning}
\nomenclature[A, 06]{DQN}{Multi layer perceptron}
\nomenclature[A, 07]{GAT}{Graph Attention Network}
\nomenclature[A, 08]{RGCN}{Relational Graph Convolutional Network}
\nomenclature[A, 09]{RDRL}{Relational Deep reinforcement learning}
\nomenclature[A, 10]{MTDRL}{Multi task deep reinforcement learning}
\nomenclature[A, 11]{CTDE}{Centralized training and decentralized execution}
\nomenclature[A, 12]{POMDP}{partially observed Markov decision process }
\nomenclature[A, 13]{PPO}{Proximal policy optimization}
\nomenclature[A, 14]{TRPO}{Trust region policy optimization}
\nomenclature[A, 15]{DDPG}{Deep deterministic policy gradient}
\nomenclature[A, 16]{GCN}{Graph convolutional network}
\nomenclature[A, 17]{NAS}{Neural Architecture Search}
\nomenclature[A, 18]{NIPA}{Node Injection poisoning attack}
\nomenclature[A, 19]{NIPA}{Node Injection poisoning attack}
\nomenclature[A, 20]{MADRL}{Multi-agent deep reinforcement learning}
\nomenclature[A, 21]{CO}{Combinatorial optimization}
\nomenclature[A, 22]{S2V}{Structure to vector}
\nomenclature[A, 23]{VRP}{vehicle routing problem}
\nomenclature[A, 24]{A2C}{Advantage Actor Critic}
\nomenclature[A, 25]{KG}{Knowledge Graph}
\nomenclature[A, 26]{KGC}{Knowledge Graph completion}
\nomenclature[A, 27]{GAN}{Generative adversarial network}
\nomenclature[A, 28]{XAI}{Explainable Artificial Intelligence}
\nomenclature[A, 28]{GPU}{Graphic processing unit}
\nomenclature[A, 29]{PGNN}{Position aware graph neural network}
\nomenclature[A, 30]{IaGNN}{Identity aware graph neural network}
\nomenclature[A, 31]{NLP}{Natural Language Processing}
\nomenclature[A, 32]{CV}{Computer Vision}




\bibliographystyle{IEEEtran}
\bibliography{references}

\end{document}